\documentclass[10pt]{article} 
\usepackage[accepted]{template/rlc}

\usepackage{amssymb}            
\usepackage{mathtools}          
\usepackage{mathrsfs}           
\usepackage{graphicx}           
\usepackage{caption, subcaption}         
\usepackage[space]{grffile}     
\usepackage{url}                
\usepackage{enumitem}
\usepackage{float}

\usepackage{todonotes}
\usepackage{lipsum}
\usepackage{wrapfig}
\usepackage{pbox}
\usepackage{array}

\makeatletter
\newcommand{\subalign}[1]{%
  \vcenter{%
    \Let@ \restore@math@cr \default@tag
    \baselineskip\fontdimen10 \scriptfont\tw@
    \advance\baselineskip\fontdimen12 \scriptfont\tw@
    \lineskip\thr@@\fontdimen8 \scriptfont\thr@@
    \lineskiplimit\lineskip
    \ialign{\hfil$\m@th\scriptstyle##$&$\m@th\scriptstyle{}##$\hfil\crcr
      #1\crcr
    }%
  }%
}
\makeatother

\newcommand\blfootnote[1]{%
  \begingroup
  \renewcommand\thefootnote{}\footnote{#1}%
  \addtocounter{footnote}{-1}%
  \endgroup
}

\newcommand{\E}{\mathbb{E}}

\newcommand{\mdp}{\mathcal{M}}
\newcommand{\states}{\mathcal{S}}
\newcommand{\actions}{\mathcal{A}}
\newcommand{\rew}{r}
\newcommand{\transitions}{P}
\newcommand{\distover}[1]{\Delta(#1)}

\title{Dissecting Deep RL with High Update Ratios: \\Combatting Value Divergence}

\author{%
  Marcel Hussing{\normalfont $^\dagger$}\\
  University of Pennsylvania\\
  \texttt{mhussing@seas.upenn.edu}\\
  \And
  Claas Voelcker{\normalfont $^\dagger$}\\
  University of Toronto\\
  Vector Institute, Toronto\\
  \texttt{cvoelcker@cs.toronto.edu}\\
  \AND
  Igor Gilitschenski \\
  University of Toronto \\
  Vector Institute, Toronto\\
  \And
  Amir-massoud Farahmand\\
  University of Toronto\\
  \And
  Eric Eaton\\
  University of Pennsylvania\\
}

\begin{document}

\maketitle

\begin{abstract} 
\blfootnote{\hspace{-0.4em}$\dagger$ The two first authors contributed equally to this work.}
We show that deep reinforcement learning algorithms can retain their ability to learn without resetting network parameters in settings where the number of gradient updates greatly exceeds the number of environment samples by combatting value function divergence.
Under large update-to-data ratios, a recent study by \citet{nikishin2022primacy} suggested the emergence of a {\em primacy bias}, in which agents overfit early interactions and downplay later experience, impairing their ability to learn. 
In this work, we investigate the phenomena leading to the primacy bias. 
We inspect the early stages of training that were conjectured to cause the failure to learn and find that one fundamental challenge is a long-standing acquaintance: value function divergence. 
Overinflated Q-values are found not only on out-of-distribution but also in-distribution data and can be linked to overestimation on unseen action prediction propelled by optimizer momentum.
We employ a simple unit-ball normalization that enables learning under large update ratios, show its efficacy on the widely used  \textsf{dm\_control} suite, and obtain strong performance on the challenging dog tasks, competitive with model-based approaches. Our results question, in parts, the prior explanation for sub-optimal learning due to overfitting early data. 

\end{abstract}

\section{Introduction} \label{sec:intro}

To improve sample efficiency, contemporary work in off-policy deep reinforcement learning (RL) has begun increasing the number of gradient updates per collected environment step~\citep{janner2019mbpo,fedus2020revisiting,chen2021randomized, hiraoka2022dropout, nikishin2022primacy, doro2023barrier, schwarzer2023bigger, kim2023resetensemble}.  
As this update-to-data (UTD) ratio increases, various novel challenges arise.
Notably, a recent study proposed the emergence of a \emph{primacy bias} in deep actor critic algorithms, defined as ``a tendency to overfit initial experiences that damages the rest of the learning process''~\citep{nikishin2022primacy}. 
This is a fairly broad explanation of the phenomenon, leaving room for investigation into how fitting early experiences causes suboptimal learning behavior.

First approaches to tackle the learning failure challenges have been suggested, such as completely resetting networks periodically during the training process and then retraining them using the contents of the replay buffer~\citep{nikishin2022primacy, doro2023barrier}. 
Resetting network parameters is a useful technique in that, in some sense, it can circumvent any previous optimization failures without prior specification. 
Yet it seems likely that a more nuanced treatment of the various optimization challenges in deep RL might lead to more efficient training down the line. 
Especially if the objective is efficiency, throwing away all learned parameters and starting from scratch periodically is counter-productive, for instance in scenarios where, keeping all previous experience is infeasible. 
As such, we set out to study the components of early training that impair learning more closely and examine whether high-UTD learning without resetting is possible. 

To motivate our study, we repeat the priming experiment of \citet{nikishin2022primacy}, in which a network is updated for a large number of gradient steps on limited data. We show that during priming stages of training, value estimates diverge so far---and become so extreme---that it takes very long to unlearn them using new, counter-factual 
data. However, contrary to prior work, we find that it is not \emph{impossible} to learn even after priming, it merely takes a long time and many samples. 
This sparks hope for our endeavor of smooth learning in high-UTD regimes.
We show that compensating for the value function divergence allows learning to proceed. This suggests that the failure to learn does not stem from overfitting early data, which would result in correct value function on seen data, but rather from improperly fitting Q-values. 
We demonstrate that this divergence is most likely caused by prediction of out-of-distribution (OOD) actions that trigger large gradient updates, compounded by the momentum terms in the Adam optimizer~\citep{kingma2015adam}.

The identified behavior, although triggered by OOD action prediction, seems to be more than the well-known overestimation due to statistical bias \citep{thrun1993issues}. 
Instead, we hypothesize that the problem is an optimization failure and focus on limiting the exploding gradients from the optimizer via architectural changes.
The main evidence for this hypothesis is that standard RL approaches to mitigating bias, such as minimization over two independent critic estimates~\citep{fujimoto2018addressing}, are insufficient. In addition, using pessimistic updates~\citep{fujimoto2019bcq, fujimoto2021td3bc} or regularization~\citep{krogh1991simple, srivastava14dropout} to treat the value divergence can potentially lead to suboptimal learning behavior, which is why architectural improvements are preferable in many cases.

We use a simple feature normalization method \citep{zhang2019root, wang2020striving, bjorck2022is} that projects features onto the unit sphere.
This decouples learning the scale of the values from the first layers of the network and moves it to the last linear layer.
Empirically, this approach fully mitigates diverging Q-values in the priming experiment. Even after a large amount of priming steps, the agent immediately starts to learn. 
In a set of experiments on the \textsf{dm\_control} MuJoCo benchmarks~\citep{tunyasuvunakool2020dmcontrol}, we show that accounting for value divergence can achieve significant across-task performance improvements when using high update ratios. 
Moreover, we achieve non-trivial performance on the challenging dog tasks that are often only tackled using model-based approaches. We demonstrate comparable performance with the recently developed TD-MPC2~\citep{hansen2024tdmpc}, without using models or advanced policy search methods.
Lastly, we isolate more independent failure modes, giving pointers towards their origins. In Appendix~\ref{app:open} we list open problems whose solutions might illuminate other RL optimization issues.

\section{Preliminaries} \label{sec:preliminaries}

{\bf Reinforcement learning}~~ \label{sec:rl} 
We phrase the RL problem~\citep{sutton2018introduction} via the common framework of solving a discounted Markov decision process (MDP)~\citep{puterman1994markov} $\mdp = \{\states, \actions, \transitions, \rew, \gamma\}$. Here, $\states$ denotes the state space, $\actions$ the action space, $\transitions(s' | s, a)$ the transition probabilities  when executing action $a$ in state $s$, $\rew(s, a)$ the reward function, and $\gamma$ the discount factor. A policy $\pi$ encodes a behavioral plan in an MDP via a mapping from states to a distribution over actions $\pi: \states \rightarrow \distover{\actions}$. 
The goal is to find an optimal policy $\pi^{*}$ that maximizes the sum of discounted return
$J_t = \sum_{k=t+1}^{\infty} \gamma^{k-t-1} r_k(s, a)$. The value function $V_{\pi}(s) = \E_{\pi, \transitions}[J_t \mid s_t = s]$ and the Q-value function $Q_{\pi}(s, a) = \E_{\pi, \transitions}[J_t \mid s_t = s, a_t = a]$ define the expected, discounted cumulative return given that an agent starts in state $s_t$ or starts in state $s_t$ with action $a_t$ respectively.

{\bf Deep actor-critic methods}~~ \label{sec:dac}
We focus on the setting of deep RL with off-policy actor-critic frameworks for continuous control~\citep{lillicrap2016ddpg, fujimoto2018addressing, haarnoja2018sac}. Our analysis uses the soft-actor critic (SAC) algorithm~\citep{haarnoja2018sac}, but our findings extend to other methods such as TD3~\citep{fujimoto2018addressing}. Commonly used off-policy actor critic algorithms like SAC have four main components: a policy $\pi_{\psi}(a|s)$, a critic network $Q_{\theta}(s,a)$, a delayed target network $\bar{Q}_{\bar{\theta}}(s,a)$, and a replay buffer $\mathcal{D} = \{s_i, a_i, r_i, s_{i+1}\}_{i=1}^N$ that stores past interaction data.
All functions are parameterized as neural networks (by $\psi$, $\theta$, and $\bar{\theta}$, respectively) and, except for the target network, updated via gradient descent. The target network is updated using Polyak averaging~\citep{polyak1992acceleration} at every time-step, formulated as $\bar{\theta} \leftarrow (1 - \tau) \bar{\theta} + \tau \theta$, where $\tau$ modulates the update amount. Actor and critic are updated using the objectives
\begin{align}
    &\!\!\max_{\psi}~\!  \mathbb{E}_{\subalign{s \sim \mathcal{D} \\ a \sim \pi_\psi(\cdot|s_{i})}}\left[\min_{j \in \{1,2\}}\!\!Q_{\theta_j}(s, a)\right],\label{eq:policy_update} \\
    &\min_{\theta} \! \left(\!Q_\theta(s, a) - \left(\!r + \gamma \mathbb{E}_{a' \sim \pi_{\psi}(\cdot|s_{i+1})}\biggl[\min_{j \in \{1,2\}}\!\!\bar{Q}_{\bar{\theta}_j}(s'\!, a')\biggr]\right)\!\! \right)^{\!2} ,
\end{align} 
respectively. In SAC, the update rules additionally contain a  regularization term that maximizes the entropy of the actor $H\left(\pi_{\psi}(\cdot) | s\right)$. 
The differentiability of the expectation in Equation~\eqref{eq:policy_update} is ensured by choosing the policy from a reparameterizable class of density functions
\citep{haarnoja2018sac}. 
We assume that all Q-functions consist of a multi-layer perceptron (MLP) encoder $\phi$ and a linear mapping $w$ such that $Q_{\theta}(s, a) = \phi(s, a) w$, where we omit parametrization of the encoder for brevity. 

\section{Investigating the effects of large update-to-data ratios during priming} \label{sec:investigating}

\begin{figure}[t!]
\centering
    \begin{subfigure}[b]{0.8\textwidth}
        \centering
        \includegraphics[height=0.4cm]{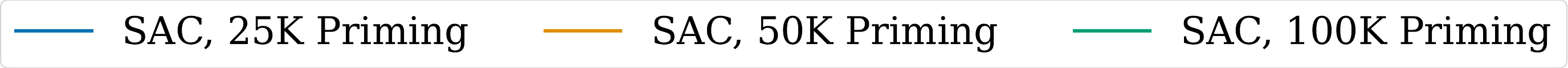}
    \end{subfigure}\\%
    \begin{subfigure}[b]{0.25\textwidth}
        \centering
        \includegraphics[width=3.7cm, trim=1cm 1cm 1cm 1cm ,clip]{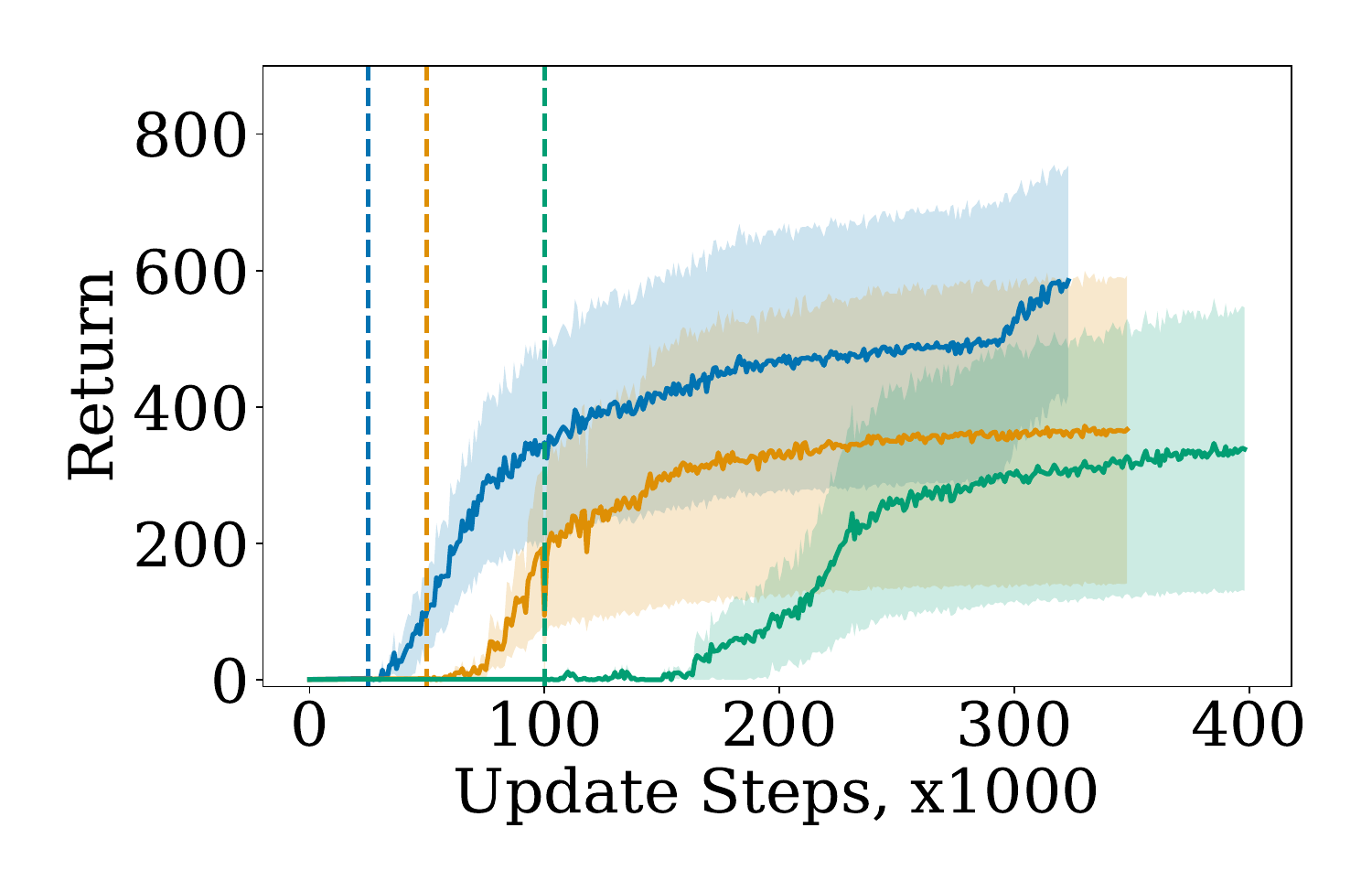}
        \label{subfig:priming_base_ret}
    \end{subfigure}%
    \begin{subfigure}[b]{0.25\textwidth}
    \centering
        \includegraphics[width=3.7cm, trim=1cm 1cm 1cm 1cm ,clip]{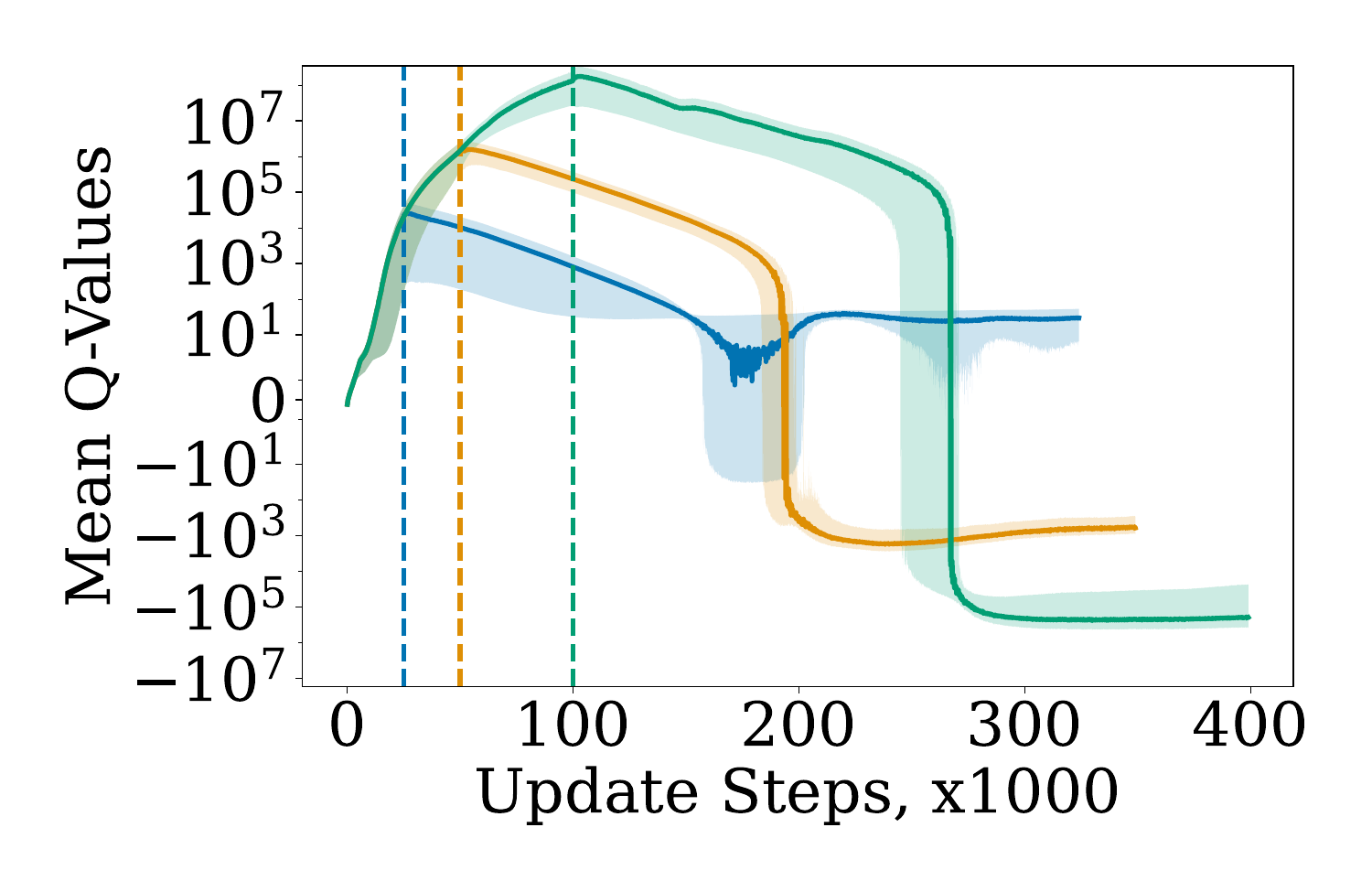}
        \label{subfig:priming_base_Q}
    \end{subfigure}%
    \begin{subfigure}[b]{0.25\textwidth}
        \centering
        \includegraphics[width=3.7cm, trim=1cm 1cm 1cm 1cm ,clip]{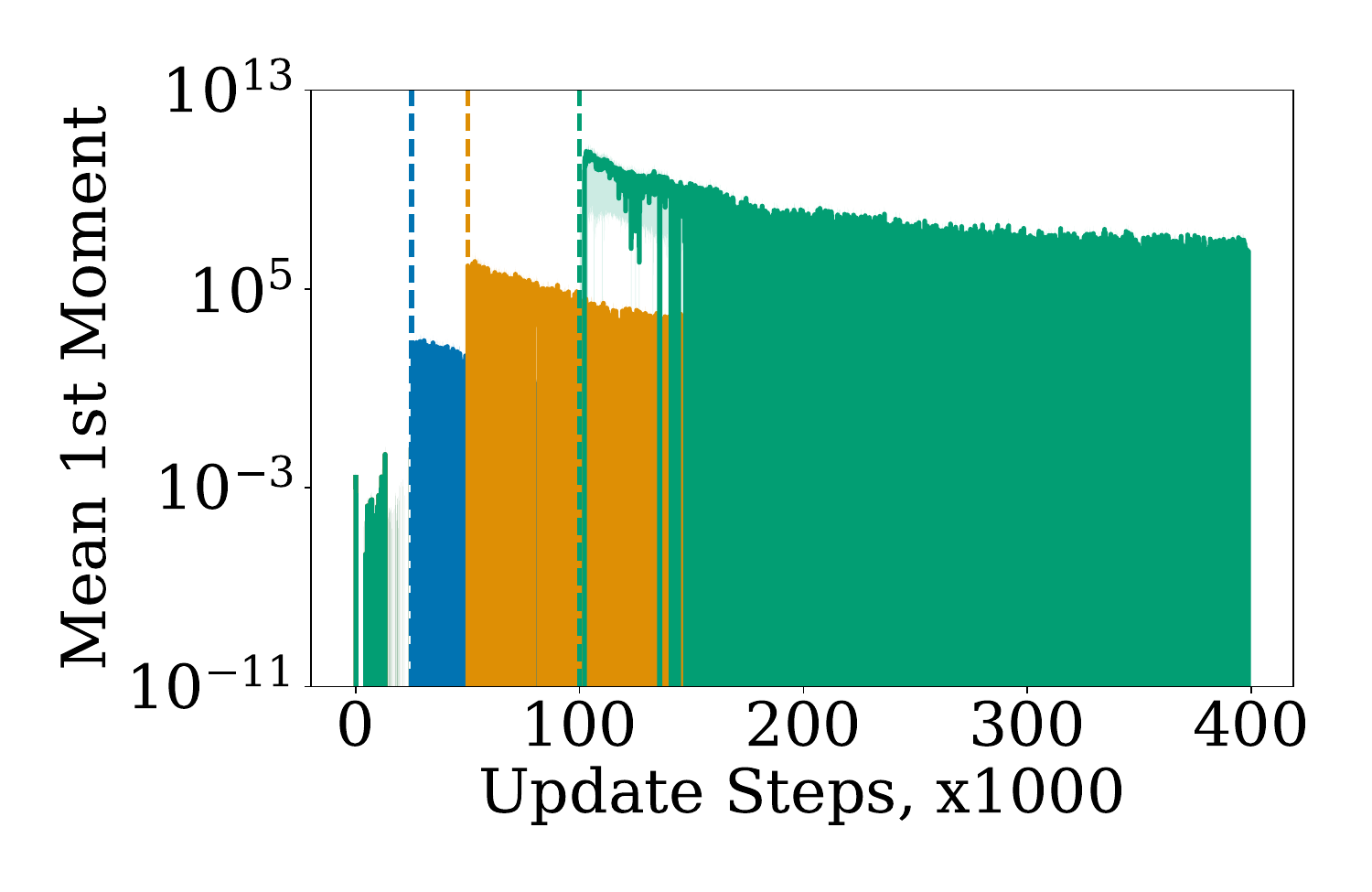}
        \label{subfig:priming_base_mom}
    \end{subfigure}%
    \begin{subfigure}[b]{0.25\textwidth}
        \centering
        \includegraphics[width=3.7cm, trim=1cm 1cm 1cm 1cm ,clip]{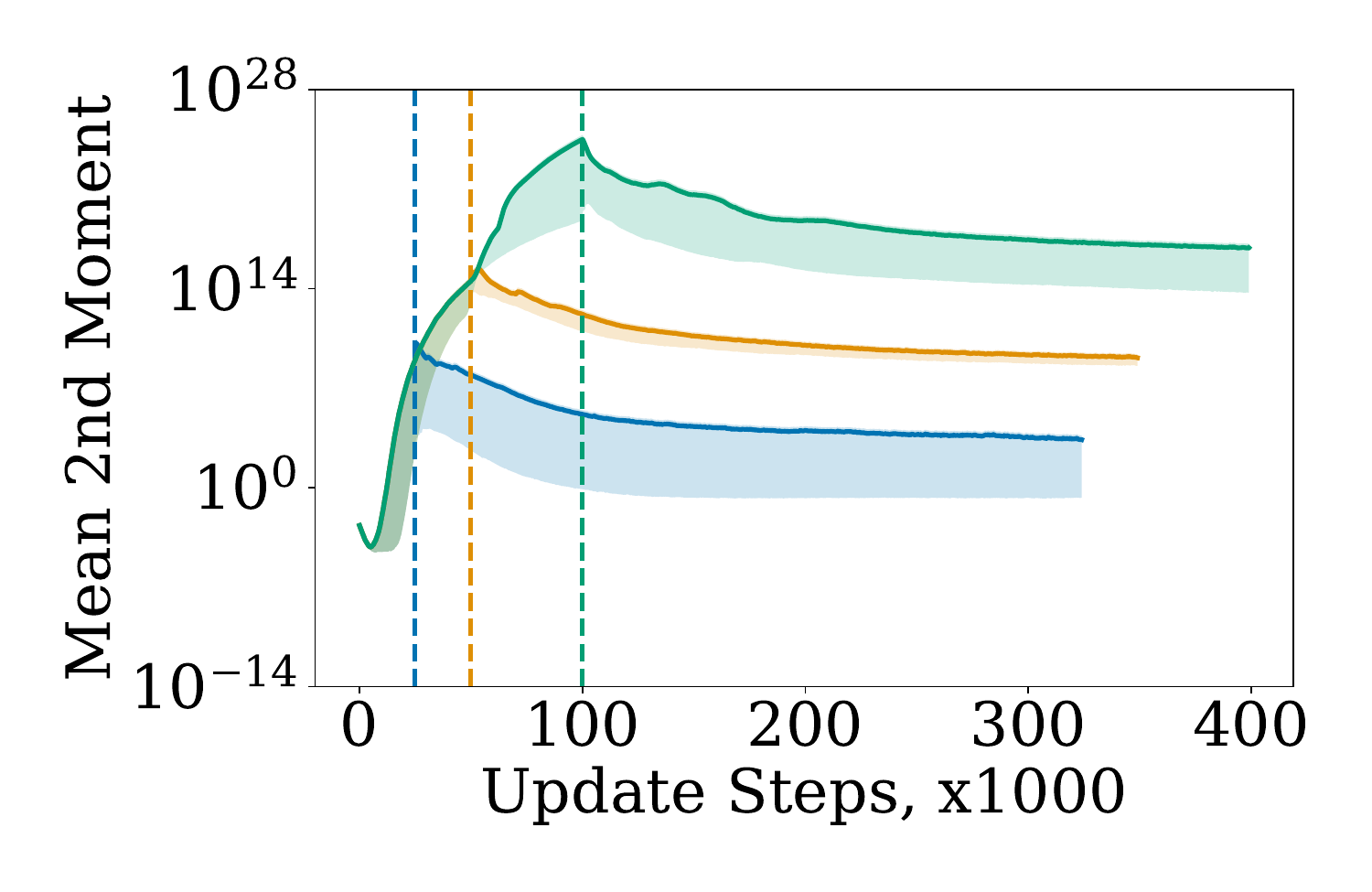}
        \label{subfig:priming_base_mom}
    \end{subfigure}%
    \vspace{-5pt}
    \caption{Return, in-distribution Q-values and Adam optimizer moments during priming for different lengths. Dotted lines correspond to end of priming. More priming leads to lower return and larger Q-value and optimizer divergence.}
    \label{fig:priming_base}
\end{figure}

As mentioned, the definition of the primacy bias is broad.
To obtain a more nuanced understanding, we set out to re-investigate the early stages of high-UTD training. To do so, we repeat the priming experiment conducted by~\citet{nikishin2022primacy}.\footnote{For consistency with later sections, we use the ReLU activation here which can lead to unstable learning of other components. We repeat all the experiments with ELUs in Appendix~\ref{app:priming} to provide even stronger support of our findings.}
We first collect a small amount of random samples. Then, using the SAC algorithm, we perform a priming step, training the agent for a large number of updates without additional data. After priming, training continues as usual. Prior results reported by~\citet{nikishin2022primacy} suggest that once the priming step has happened, agents lose their ability to learn completely. We use the simple Finger-spin task \citep{tunyasuvunakool2020dmcontrol} to study the root causes for this systematic failure and to examine if there are ways to recover without resets. In this section, we report means over five random seeds with standard error in shaded regions. Hyperparameters are kept consistent with previous work for ease of comparison.

\subsection{An old acquaintance: Q-value overestimation} \label{sec:overestimation}
We first ask whether there is a barrier as to how many steps an agent can be primed for before it becomes unable to learn. We test this by collecting 1,000 samples and varying the number of updates during priming from 25,000 to 50,000 and 100,000. The results are presented in Figure~\ref{fig:priming_base}. 

We make two key observations. First, lower amounts of priming are correlated with higher early performance. More precisely, it seems that many runs simply take longer before they start learning as the number of priming steps increases. Second, during priming the scale of the average Q-value estimates on observed state-action pairs increases drastically. We find that the Q-values start out at a reasonable level, but as priming goes on they eventually start to diverge drastically. Once the agent estimates very large Q-values, the final performance in terms of average returns deteriorates. We also observe that the second moment of the Adam optimizer~\citep{kingma2015adam} is correlated with the divergence effect. Optimizer divergence has been observed before as a cause of plasticity loss under non-stationarity~\citep{lyle2023understanding}, but in our experiments the data is stationary during priming. We conjecture that the momentum terms lead to much quicker propagation of poor Q-values and ultimately to prediction of incorrect Q-values, even on in-distribution data.

After priming, there exist two cases: 1) either the Q-values need to be unlearned before the agent can make progress or 2) there is a large drop from very high to very low Q-values that is strongly correlated with loss in effective dimension of the embedding, as defined by~\cite{yang2020harnessing} (see Appendix~\ref{app:priming_dim}). In the second case, rank can sometimes be recovered upon seeing new, counter-factual data and the network continues to learn. Yet, sometimes the agent gets stuck at low effective dimension; a possible explanation for the failure to learn observed in the priming experiments of~\citet{nikishin2022primacy}. This is orthogonal to a previously studied phenomenon where target network-based updates lead to perpetually reduced effective rank~\citep{kumar2021implicit}.

\subsection{On the potential causes of divergence} \label{sec:causes}

\begin{figure}[t]
\begin{minipage}[b]{.35\textwidth}
\centering
    \begin{subfigure}[b]{\textwidth}
        \centering
        \includegraphics[height=0.8cm]{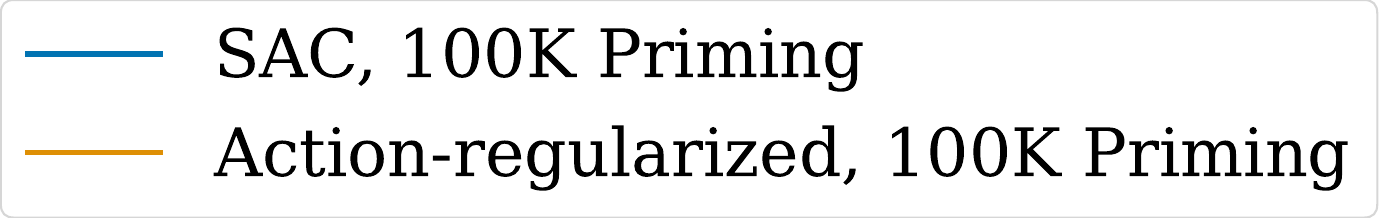}
    \end{subfigure}\\%
    \begin{subfigure}[b]{\textwidth}
        \centering
        \includegraphics[width=4.8cm, trim=1cm 1cm 1cm 1cm ,clip]{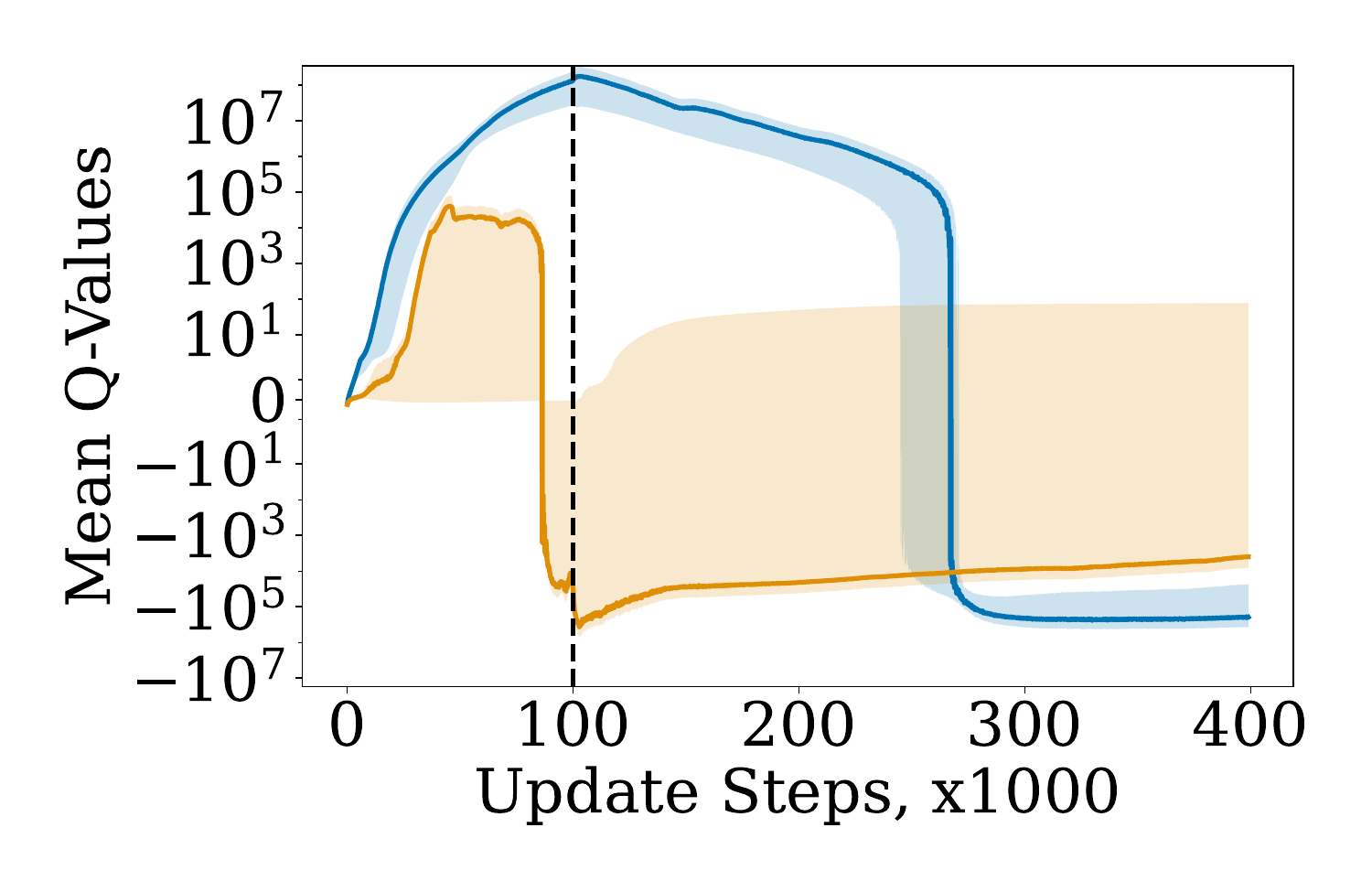}
        \label{subfig:priming_causes_Q}
    \end{subfigure}%
    \vspace{-5pt}
    \caption{Priming with SAC and action regularization during priming. The latter lowers divergence. }
    \label{fig:priming_causes}
\end{minipage}
\hfill
\begin{minipage}[b]{.62\textwidth}
    \centering
    \begin{subfigure}[b]{0.8\textwidth}
        \centering
        \includegraphics[height=0.8cm]{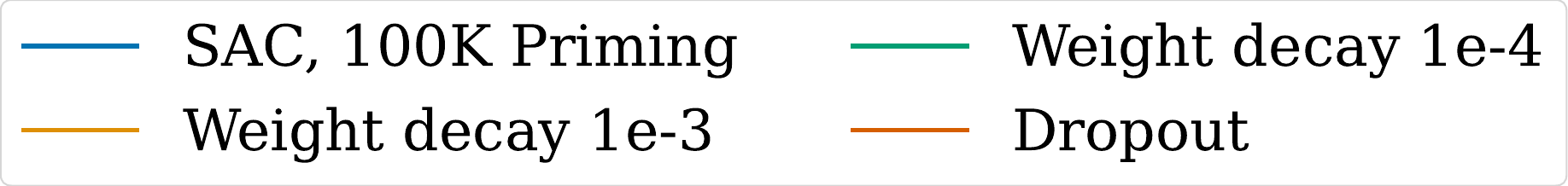}
    \end{subfigure}\\%
    \begin{subfigure}[b]{0.5\textwidth}
        \centering
        \includegraphics[width=4.8cm, trim=1cm 1cm 1cm 1cm ,clip]{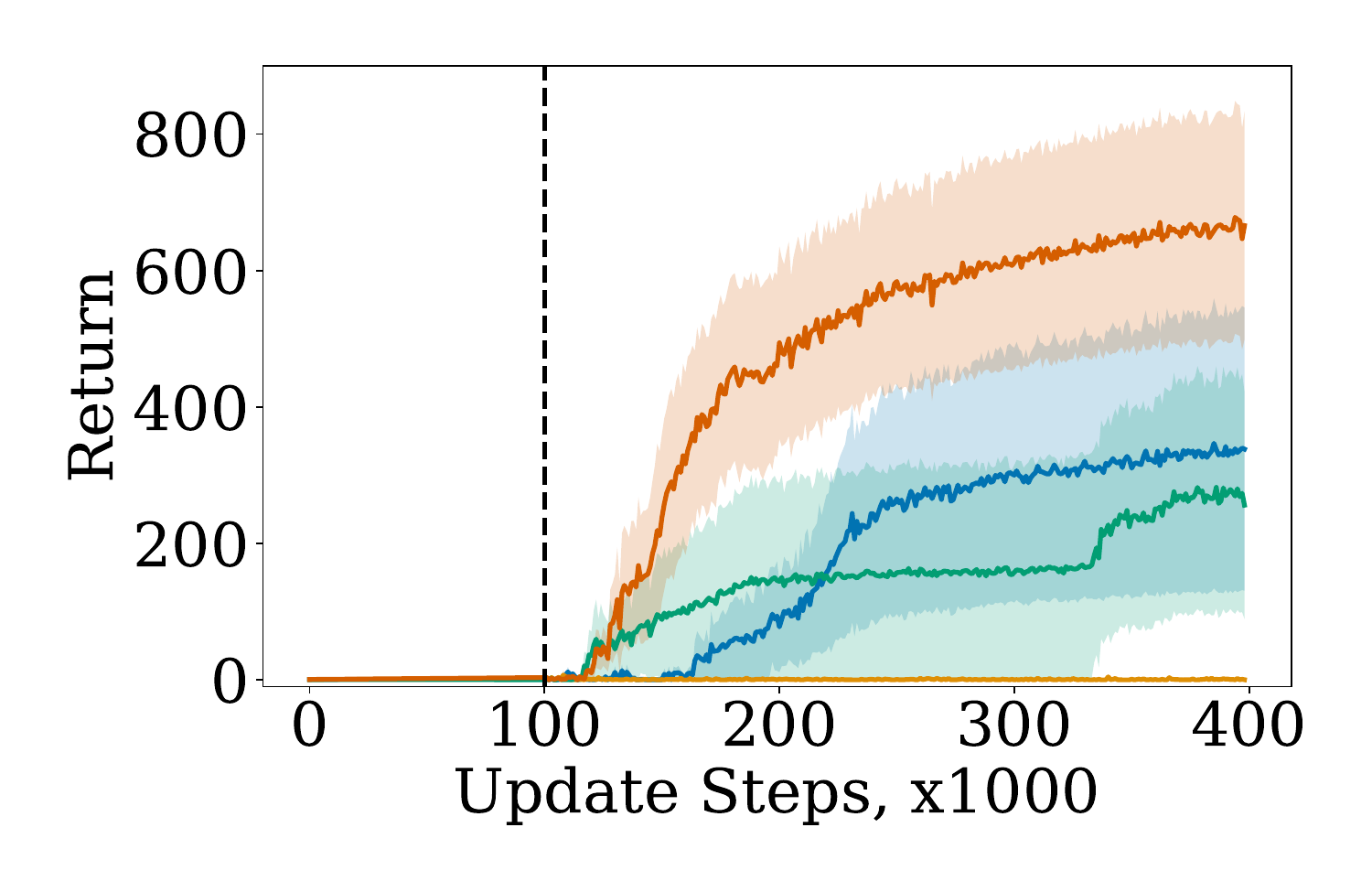}
        \label{subfig:priming_abl_ret}
    \end{subfigure}%
    \begin{subfigure}[b]{0.5\textwidth}
    \centering
        \includegraphics[width=4.8cm, trim=1cm 1cm 1cm 1cm ,clip]{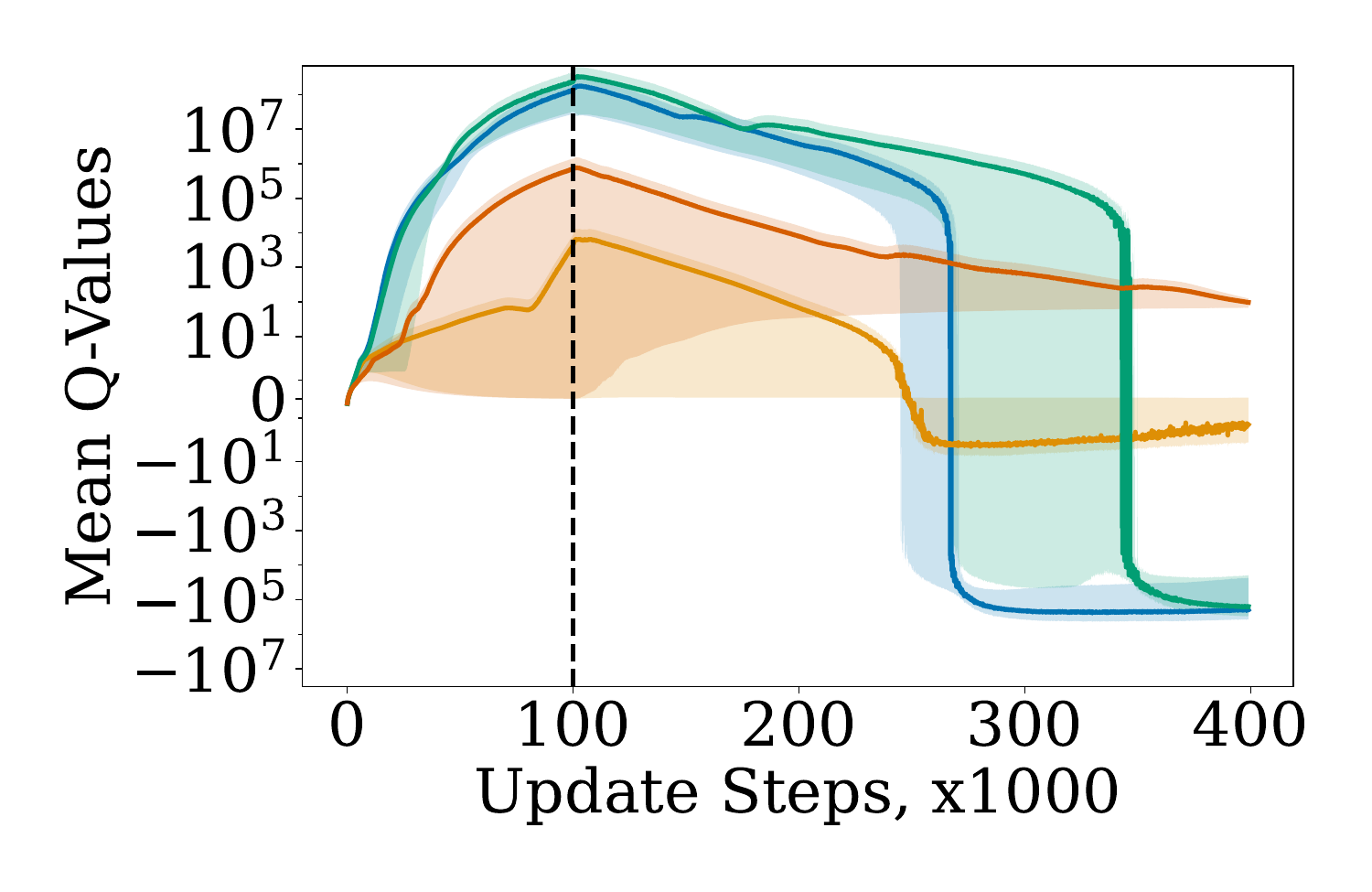}
        \label{subfig:priming_abl_Q}
    \end{subfigure}%
    \vspace{-18pt}
    \caption{Return and
    Q-values of priming runs with weight decay and dropout. Results indicate that both regularizations mitigate priming to some extent but not sufficiently. }
    \label{fig:priming_abl}
\end{minipage}
    \vspace{-5pt}
\end{figure}

We conjecture that Q-value divergence starts with overestimated values of OOD actions. This overestimation could cause the optimizer to continually increase Q-values via its momentum leading to divergence. 
To test this hypothesis, we add a conservative behavioral cloning~\citep{pomerleau1988alvinn, atkeson1997robot} loss term to our actor that forces the policy to be close to replay buffer actions. Prior work employed this technique in offline RL to mitigate value overestimation~\citep{fujimoto2021td3bc}. More formally, our actor update is extended by the loss $
        \mathcal{L}_{c, \psi} = \min_{\psi} \mathbb{E}_{a \sim \mathcal{D}, \hat{a} \sim \pi_{\psi}(s)}[||a - \hat{a}||_2]$. 
The results in Figure~\ref{fig:priming_causes} indicate that the basis of the conjecture is corroborated as divergence is much smaller---but not mitigated completely---when values are learned on actions similar to seen ones. However, in practice we do not know when divergence sets in, which limits the applicability of this technique in realistic scenarios. Using it throughout all of training, rather than just during priming, impairs the learner's ability to explore. We investigate the effects of the optimizer in more detail and provide preliminary evidence that the second-order term may be at fault in Appendix~\ref{app:priming_opt}. 

\subsection{Applying common regularization techniques} \label{sec:regularization}

Regularization is a common way to mitigate gradient explosion and is  often used to address overestimation~\citep{farebrother2018generalization, chen2021randomized, liu2021regularization, hiraoka2022dropout, li2023efficient}. We investigate the priming experiments under techniques such as using L$^2$ weight decay~\citep{krogh1991simple} or adding dropout~\citep{srivastava14dropout} to our networks in Figure~\ref{fig:priming_abl}.

Both L$^2$ weight decay as well as dropout can somewhat reduce the divergence during priming, however not to a sufficient degree.
While L$^2$ regularization fails to attain very high performance, dropout is able to recover a good amount of final return.
However, both methods require tuning of a hyperparameter that trades off the regularization term with the main loss.
This hyperparameter is environment-dependent and tuning it becomes infeasible for large UTD-ratios due to computational resource limitations. 
Still, the results imply that it is in fact possible to overcome the divergence in priming and continue to learn good policies.

\subsection{Divergence in practice} 
\label{sec:real}

\begin{wrapfigure}{r}{0.54\textwidth}
\vspace{-5pt}
  \begin{minipage}{\linewidth}
    \centering\captionsetup[subfigure]{justification=centering}
    \includegraphics[width=4cm, trim=1cm 1cm 1cm 1cm ,clip]{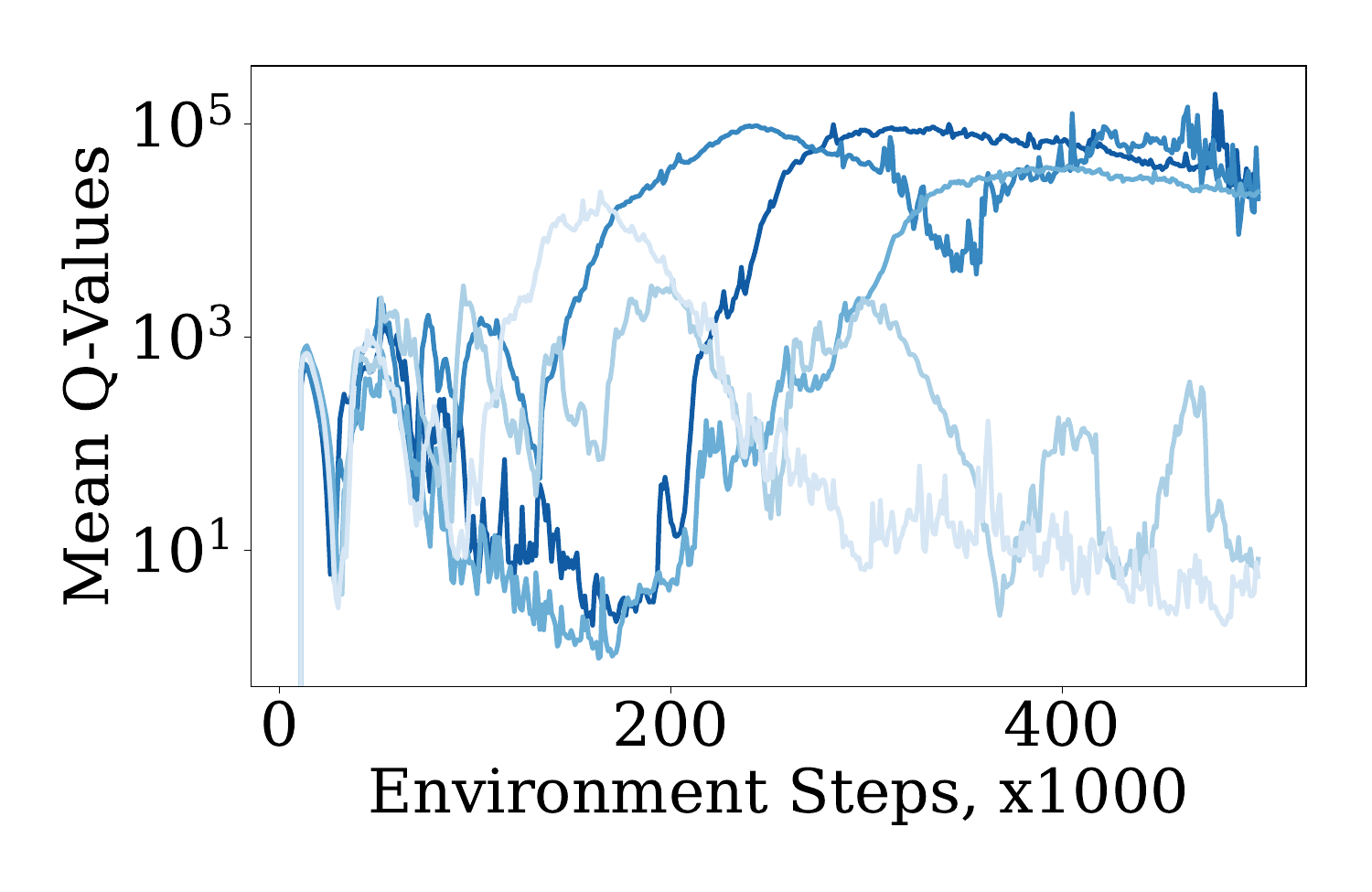}\vspace{-8pt}
    \includegraphics[width=4cm, trim=1cm 1cm 1cm 1cm ,clip]{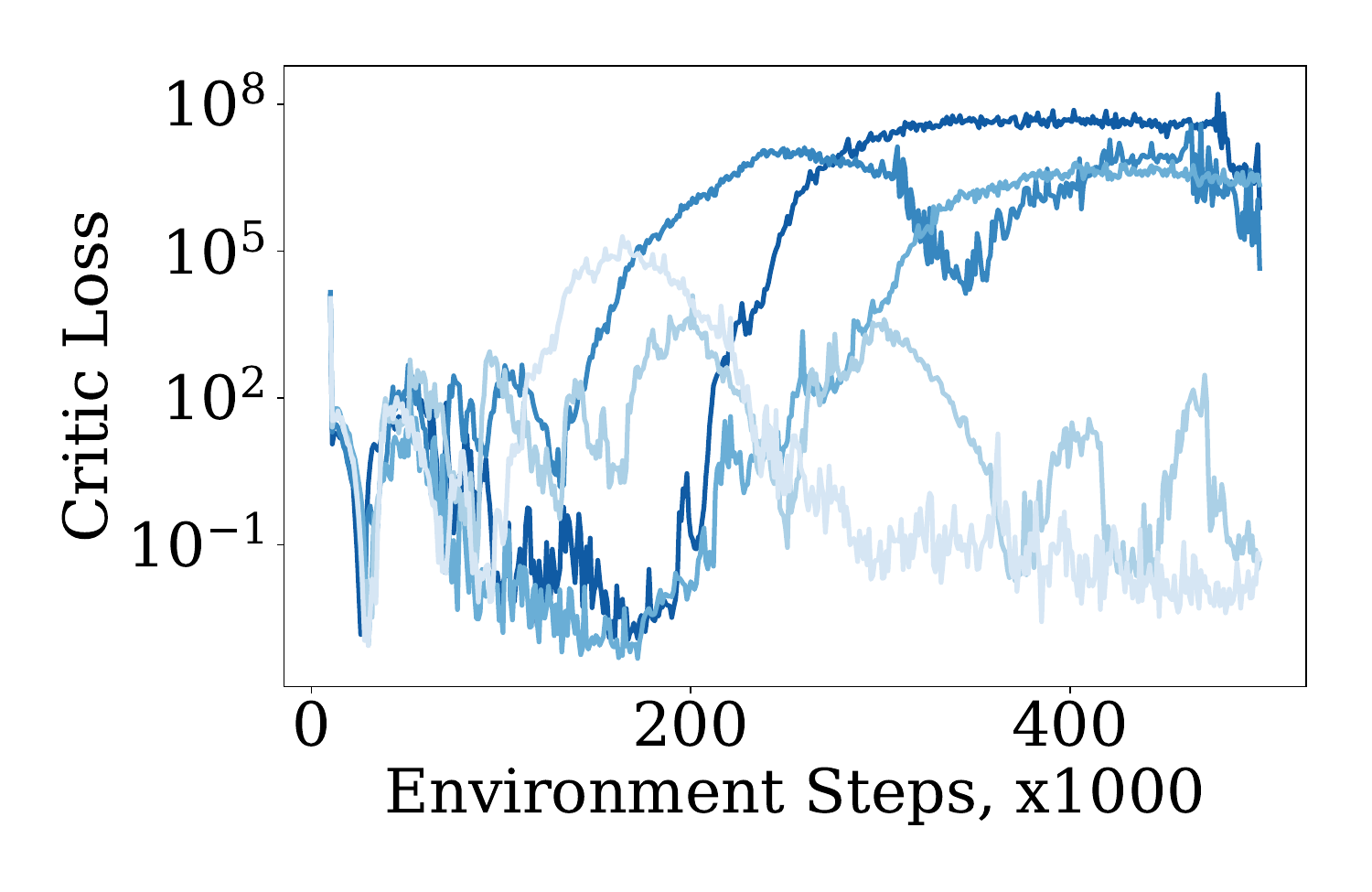}
    \caption{In-distribution Q-values and critic loss of five SAC seeds on the humanoid-run task using $\mathrm{UTD}=32$. Values diverge at arbitrary time-points, not only during the beginning. Loss mirrors Q-value divergence.}
    \label{fig:humanoid_failure}
  \end{minipage}
  \vspace{-15pt}
\end{wrapfigure}

One question that remains is whether we can find these divergence effects outside of the priming setup. We find that, while priming is an artificially constructed worst case, similar phenomena happen in regular training on standard benchmarks when increasing update ratios (see Figure~\ref{fig:humanoid_failure}). Further, the divergence is not limited to early stages of training as it happens at arbitrary points in time.
We therefore conjecture that divergence is not a function of 
amount of experience but rather one of state-action space coverage. Note that the reported Q-values have been measured on the observed training data, not on any out-of-distribution state-action pairs. 
The respective critic losses become very large. All this points toward a failure to fit Q-values. This behavior does not align with our common understanding of overfitting~\citep{bishop2006pattern},  challenging the hypothesis that high-UTD learning fails merely due to large validation error~\citep{li2023efficient}.

\section{Towards high-UTD optimization without resetting} \label{sec:method}

Regularization techniques such as those in Section~\ref{sec:regularization} can fail to alleviate divergence as they tend to operate across the whole network and lower the weights everywhere even if higher values are actually indicated by the data. They also require costly hyperparameter tuning. Thus, we turn towards network architecture changes to the commonly used MLPs that have proven useful in overcoming issues such as exploding gradients in other domains~\citep{ba2016layer, xu2019understanding}. 

\subsection{Limiting gradient explosion via unit ball normalization} \label{sec:unitball}

\begin{figure}[t]
\begin{minipage}[t]{.60\textwidth}
\centering
    \begin{subfigure}[b]{\textwidth}
        \centering
        \includegraphics[height=0.4cm]{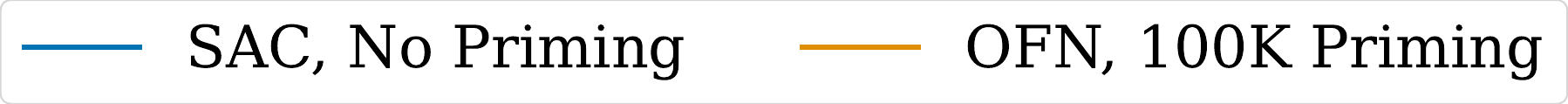}
    \end{subfigure}\\%
    \begin{subfigure}[b]{0.5\textwidth}
        \centering
        \includegraphics[width=4.8cm,clip,trim=1cm 1cm 1cm 1cm]{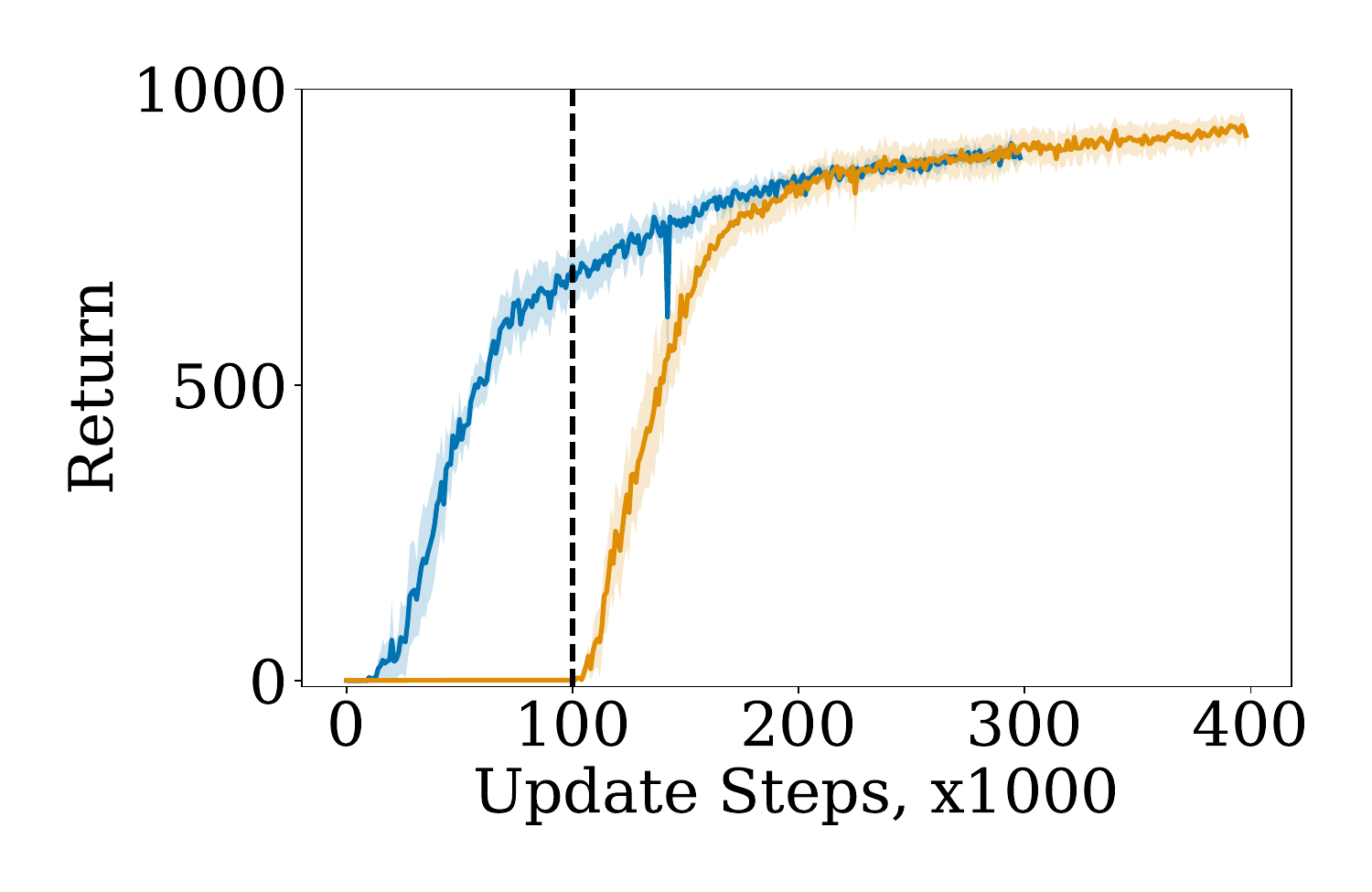}
        \label{subfig:priming_norm_ret}
    \end{subfigure}%
    \begin{subfigure}[b]{0.5\textwidth}
        \centering
        \includegraphics[width=4.8cm,clip,trim=1cm 1cm 1cm 1cm]{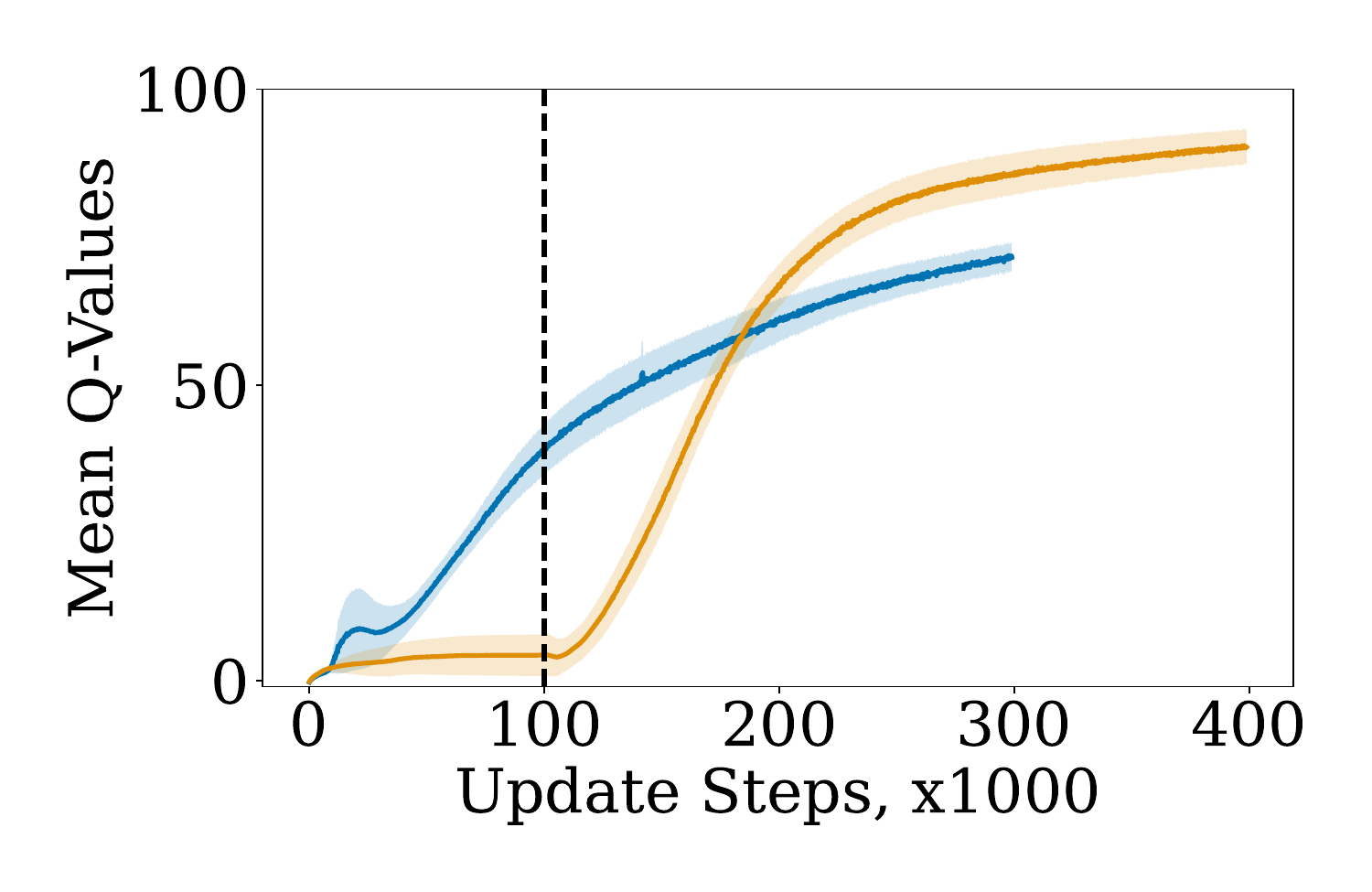}
        \label{subfig:priming_norm_Q}
    \end{subfigure}%
    \vspace{-20pt}
    \caption{(Left) Return and (Right) Q-values comparing SGD result and OFN when priming for 100K steps. OFN obtains returns close to that of the well-trained SGD agent and learns an appropriate Q-value scale correctly.}
    \label{fig:priming_norm}
\end{minipage}
\hfill
\begin{minipage}[t]{.36\textwidth}
\centering
    \begin{subfigure}[b]{\textwidth}
        \hspace{15pt}
        \centering
        \includegraphics[height=0.8cm]{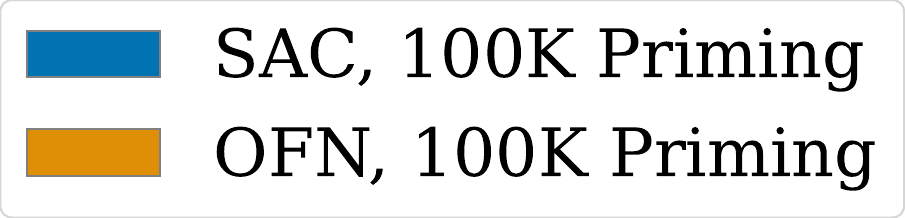}
    \end{subfigure}
    \\%
    \begin{subfigure}[b]{\textwidth}
        \centering
        \includegraphics[width=4.6cm,clip,trim=1cm 1cm 1cm 1cm]{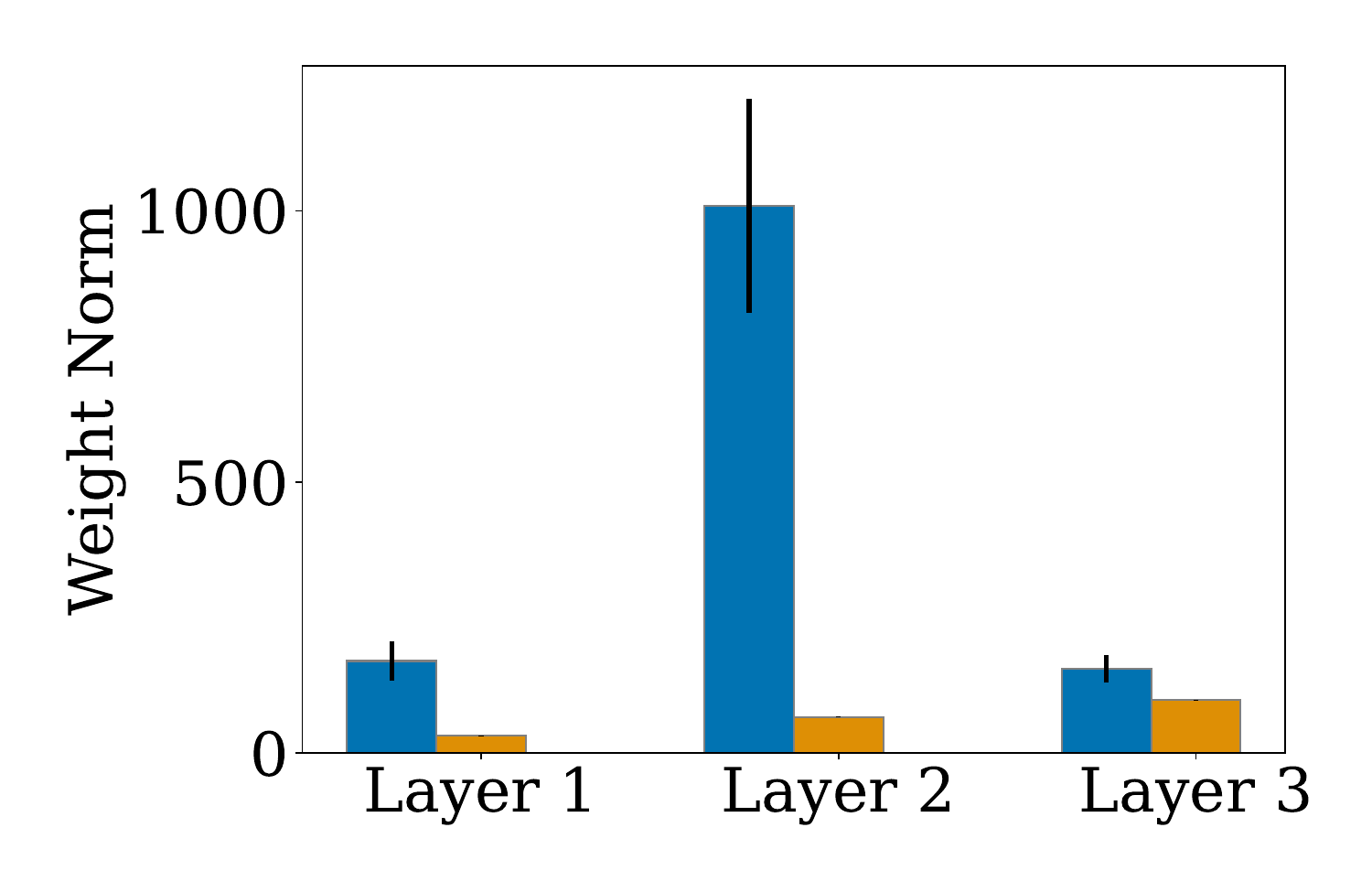}
    \end{subfigure}%
    \vspace{-4.5pt}
    \caption{$L_2$ norm of network weights per layer after priming for default and OFN  architectures. OFN leads to smaller weights and significant mass in the last layer.}
    \label{fig:weight_norm}
\end{minipage}
\vspace{-5pt}
\end{figure}

As discussed previously, the prediction of an unknown action might trigger the propagation of a large, harmful gradient. Further, the Q-values of our network ought to grow over time as they more closely approximate those of a good policy. If we predict incorrectly on one of these Q-values, a potentially very large loss is propagated. Gradients are magnified by multiplicative backpropagation via ReLU activations~\citep{glorot2011deep} as well as momentum from Adam~\citep{kingma2015adam}. Note that all resulting issues arise in the early network layers. 
We hypothesize that we can address many of these problems by separating the scaling of the Q-values to the appropriate size from the earlier non-linear layers of the network and moving the Q-value scaling to the final linear layer.

One contender to achieve the value decoupling described in the previous paragraph is layer normalization~\citep{ba2016layer}, but one would have to disable scaling factors used in common implementations. Still, standard layer normalization would not guarantee small features everywhere. Instead, we use a stronger constraint and project the output features of the critic encoder onto the unit ball using the function
$f(\mathbf{x}) = \frac{\mathbf{x} }{\|\mathbf{x}\|_2}$~\citep{zhang2019root}, 
where $\|\cdot\|_2$ denotes the L$^2$ vector norm and $\mathbf{x}$ is the output of our encoder $\phi(s, a)$. This ensures that all values are strictly between $0$ and $1$ and the gradients will be tangent to the unit sphere. Note that this function's gradient is not necessarily bounded to ensure low gradient propagation (see Appendix~\ref{app:unitnorm}), but we argue that if large values are never created in the early layers, gradient explosion will not occur. The unit ball has previously been used to mitigate large action prediction in the actor~\citep{wang2020striving} or to stabilize RL training in general~\citep{bjorck2022is}. 
For brevity, we will refer to this method as {\em output feature normalization} (OFN). We solely apply OFN to the critic, unlike~\citet{wang2020striving}, since our goal is to mitigate value divergence. OFN is very simple and requires only a one-line change in implementation.

\subsection{Evaluating feature output normalization during priming} \label{sec:evalmethod}

To test the efficacy of the OFN-based approach, we repeat the priming experiment in Figure~\ref{fig:priming_norm}.
We find that OFN achieves high reward and most distinctly, Q-value divergence during priming is fully mitigated. 
Note also that we are using a discount factor of $\gamma = 0.99$, returns are collected over 1,000 timesteps and rewards are in $[0, 1]$. 
We therefore expect the average Q-values to be roughly at $10\%$ of the undiscounted return which seems correct for the OFN network. 
However, more importantly, as shown in Figure~\ref{fig:weight_norm}, most of the Q-value scaling happens in the last layer.

\section{Experimental evaluation}

We evaluate our findings on the commonly used \textsf{dm\_control} suite~\citep{tunyasuvunakool2020dmcontrol}. All results are averaged over ten random seeds.\footnote{For comparison with TD-MPC2~\citep{hansen2024tdmpc} we use data provided by their implementation, which only contains three seeds. As the goal is not to rank algorithmic performance but to give intuition about the relative strengths of adapting the network architecture, we believe that this is sufficient in this case.} We report evaluation returns similar to~\citet{nikishin2022primacy}, which we record every 10,000 environment steps. We compare a standard two-layer MLP with ReLU ~\citep{nair2010rectified} activations, both with and without resetting, to the same MLP with OFN.  
The architecture is standard in many reference implementations. Architecture and the resetting protocol are taken from \citet{doro2023barrier} and hyperparameters are kept without new tuning to ensure comparability of the results. More details can be found in Appendix~\ref{app:impl}.

To understand the efficacy of output normalization on real environments under high UTD ratios, we set out to answer multiple questions that will illuminate RL optimization failures:\\
{\bf Q1:}~Can we maintain learning without resetting neural networks?\\
{\bf Q2:}~Are there other failure modes beside Q-value divergence under high UTD ratios?\\
{\bf Q3:}~When resets alone fall short, can architectural changes enable better high-UTD training?

\subsection{Feature normalization stabilizes high-UTD training} 

\begin{figure}[t]
\begin{minipage}[b]{0.67\textwidth}
\centering
    \begin{subfigure}[b]{\textwidth}
        \centering
        \includegraphics[width=10.2cm,clip,trim=0.3cm 0 0.3cm 0cm]{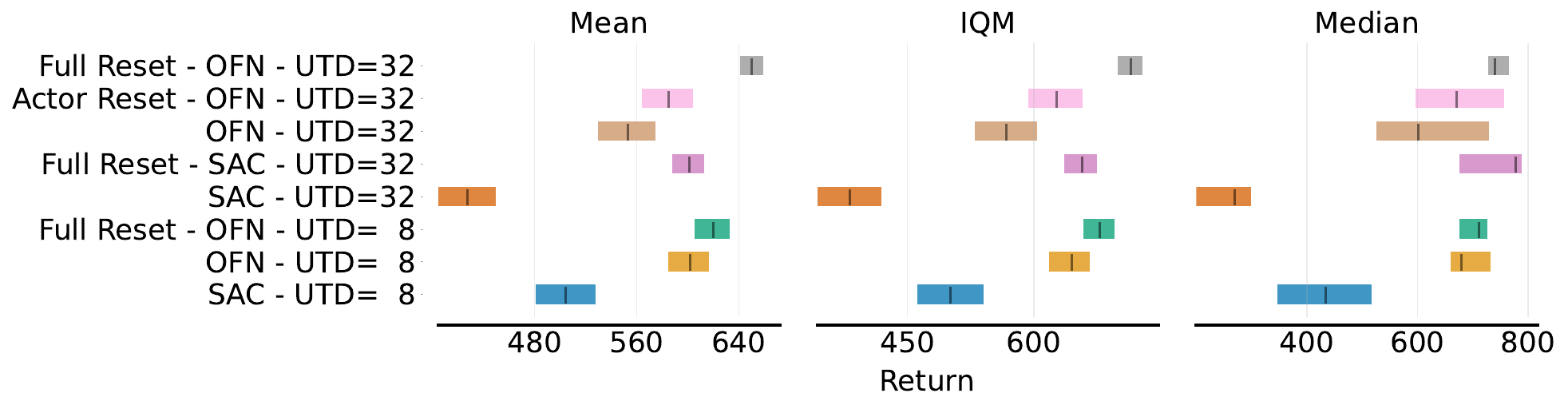}
    \end{subfigure}%
    \caption{Mean, interquartile mean (IQM), and median with $95\%$ bootstrapped confidence intervals of standard SAC and OFN on the DMC15-500k Suite. OFN can maintain high performance even under large UTD. OFN with $\mathrm{UTD} = 8$ achieves comparable performance to standard resetting with $\mathrm{UTD} = 32$ across metrics.}
    \label{fig:aggregate}
\end{minipage}
\hfill
\begin{minipage}[b]{.3\textwidth}
\centering
    \begin{subfigure}[b]{\textwidth}
        \centering
        \includegraphics[width=4.25cm,clip,trim=0.3cm 0 0 0]{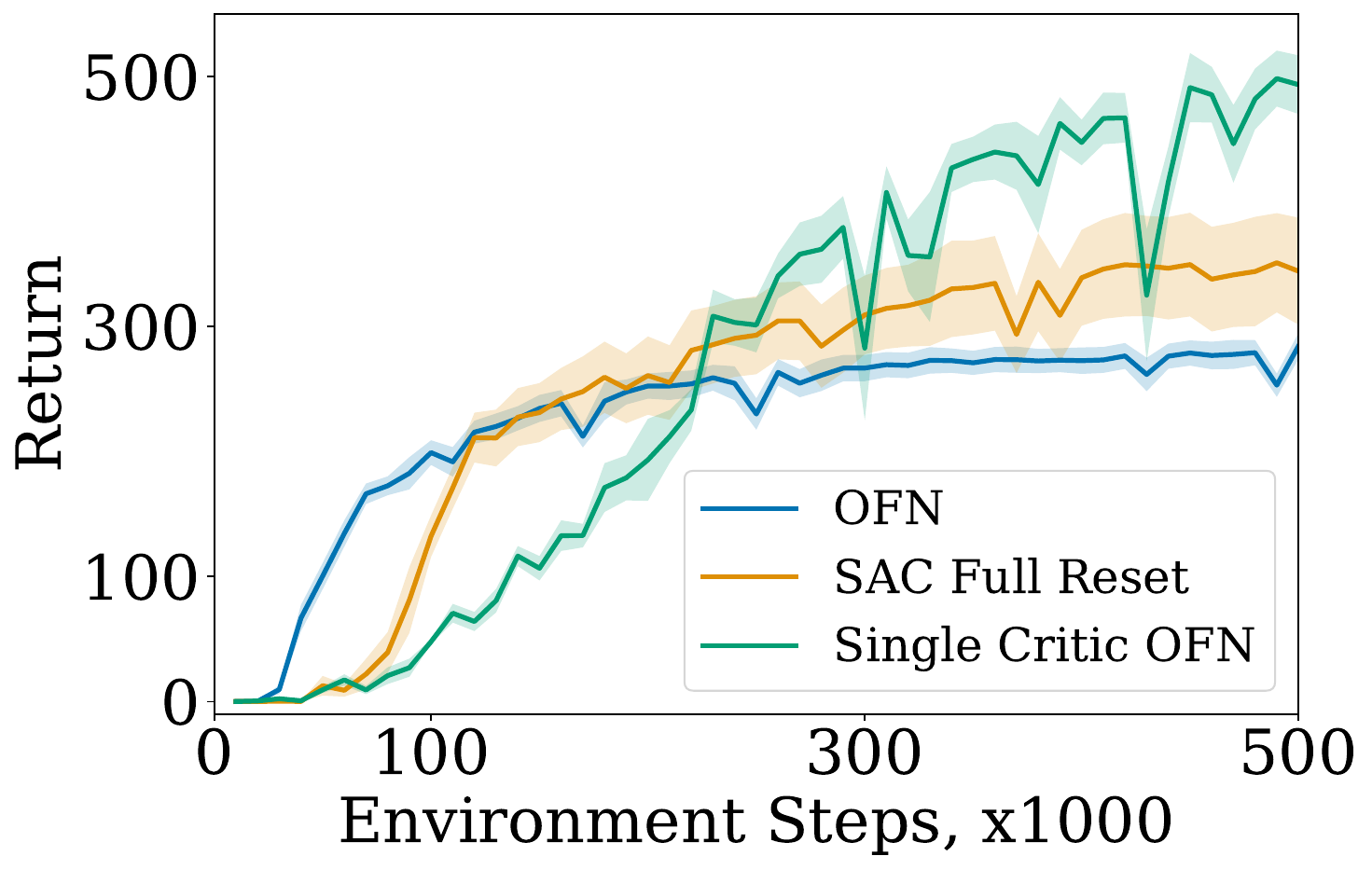}
    \end{subfigure}%
    \caption{Mean return of single-critic OFN, standard OFN and resetting; $\mathrm{UTD}=32$ on hopper-hop. Shaded regions are standard error.}
    \label{fig:hopper_hop}
\end{minipage}
\vspace{-5pt}
\end{figure}

To answer \textbf{Q1}, we compare OFN and SAC with resets on the DMC15-500k benchmark with large update ratios of $8$ and $32$ as proposed by \citet{nikishin2022primacy} and  \citet{schwarzer2023bigger}.  We report mean, interquartile mean (IQM) and median as well as $95\%$ bootstrapped confidence intervals aggregated over seeds and tasks,  following~\citet{agarwal2021deep}. The results are shown in Figure~\ref{fig:aggregate}.

First, we observe that in both cases, $\mathrm{UTD}=8$ and $\mathrm{UTD}=32$, OFN can significantly improve over the non-resetting MLP baseline across all metrics. The value estimates that diverge seem to have been handeled properly (see Appendix~\ref{app:exp_q}); learning is maintained. We note that our approach with $\mathrm{UTD}=8$ achieves mean and IQM performance comparable to that of standard resetting with $\mathrm{UTD}=32$. In median and quantile performance, all $\mathrm{UTD}=32$ overlap, highlighting that outliers contribute to the performance measurement. Note that the overall performance drops slightly for the OFN-based approach when going from $\mathrm{UTD}=8$ to $\mathrm{UTD}=32$. We conjecture that there are other learning problems such as exploration that have not been treated by alleviating value divergence. However, these do not lead to complete failure to learn but rather slightly slower convergence. 

\subsection{Other failure modes: Exploration limitations} \label{sec:otherfailure}

To validate the hypothesis of other failures and answer \textbf{Q2}, we run two additional experiments. First, our current focus is on failures of the critic; our proposed mitigation does not address any further failures that might stem from the actor. We defer a more detailed analysis of actor failure cases to future work. Instead, we test the OFN-based architecture again and, for now, simply reset the actor to shed light on the existence of potential additional challenges.
For comparison, we also run a second experiment in which we reset all learned parameters, including the critic.

The results in Figure~\ref{fig:aggregate} indicate that actor resetting can account for a meaningful portion of OFN's performance decline when going from $\mathrm{UTD}=8$ to $\mathrm{UTD}=32$. The actor-reset results are within variance of the full-resetting standard MLP baseline. Further, we observe that there is still some additional benefit to resetting the critic as well. 
This does not invalidate the premise of our hypothesis, value divergence might not be the \emph{only} cause of problems in the high UTD case. We have provided significant evidence that it is a \emph{major} contributor.
Resetting both networks of OFN with $\mathrm{UTD}=32$  outperforms all other baselines on mean and IQM comparisons. 

To explain the remaining efficacy of critic resets, we examine the hopper-hop environment where standard SAC with resets outperforms OFN. 
In RL with function approximation, one might not only encounter over-  but also under-estimation~\citep{wu2020reducing, lan2020maxmin, saglam2021estimation}.
We believe that hopper  is sensitive to pessimism, and periodically resetting the networks might partially and temporarily counteract the inherent pessimism of the dual critic setup.

To obtain evidence for this conjecture, we repeated some experiments with a single critic. As OFN handles divergence it might not require a minimization over two critics~\citep{fujimoto2018addressing}. We compare OFN using a single critic and $32$ updates per environment step to standard SAC and OFN in Figure~\ref{fig:hopper_hop}. With a single critic, OFN does not get stuck in a local minimum and outperforms full resetting. Note that this is only true in few environments, leading us to believe that the effects of high-update training are MDP-dependent.
In some environments we observe unstable learning with a single critic, which highlights that the bias countered by the double critic optimization and the overestimation from optimization we study are likely orthogonal phenomena that both need to be addressed.
Most likely, there is a difficult trade-off between optimization stability and encouraging sufficient exploration, which is an exciting avenue for future research.

\subsection{Limit-testing feature normalization}

\begin{figure}[t]
\centering
    \begin{subfigure}[b]{0.8\textwidth}
        \centering
        \includegraphics[height=0.4cm]{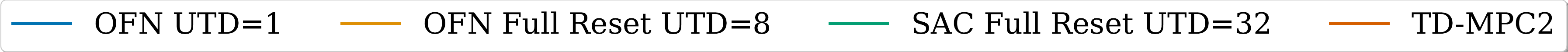}
    \end{subfigure}\\%
    \hspace{-20pt}
    \begin{subfigure}[t]{0.25\textwidth}
        \centering
        \includegraphics[width=4.cm, trim=0.4 0 0 0 ,clip]{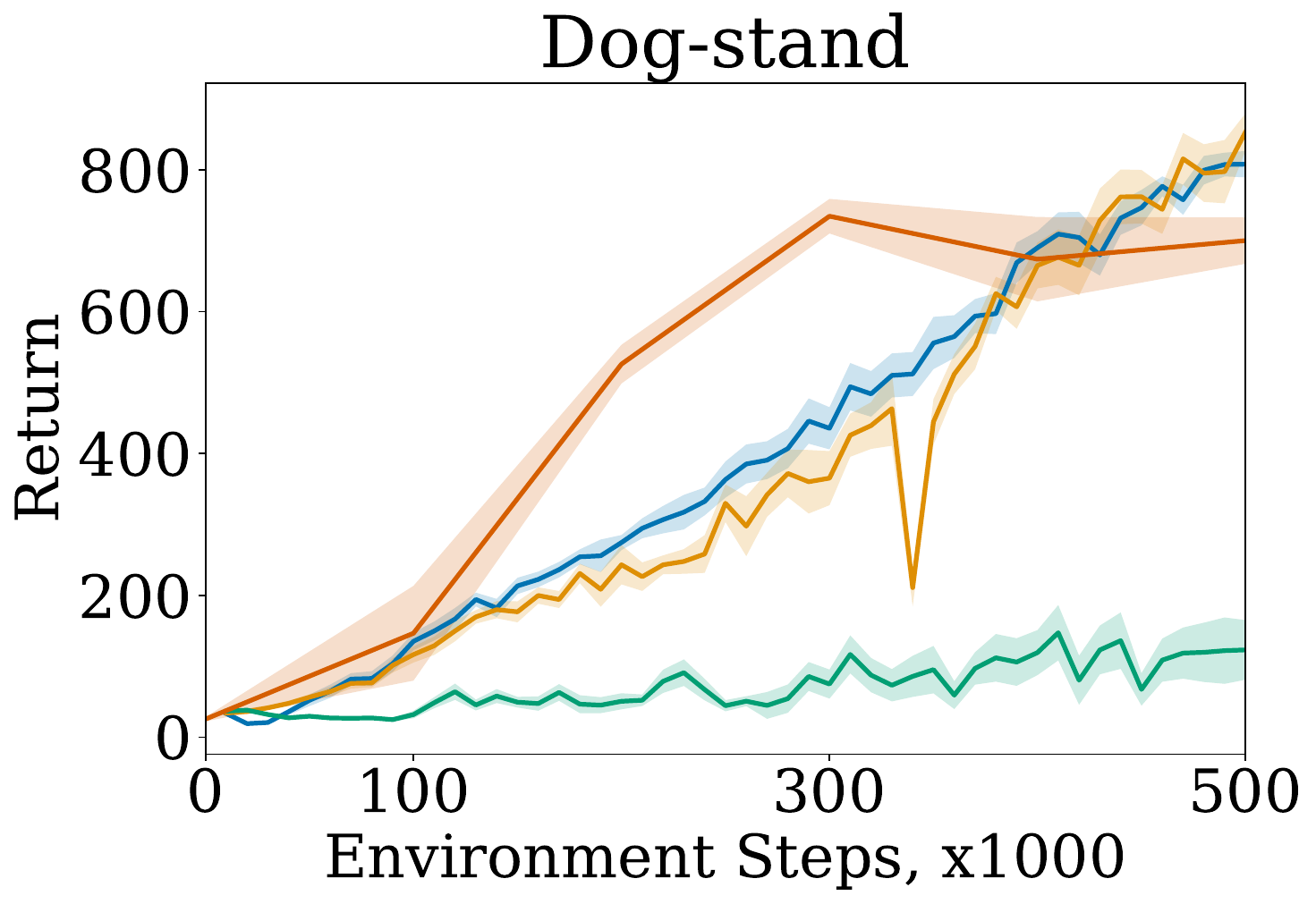}
        \label{subfig:dog_stand}
        \vspace{-12pt}
    \end{subfigure}%
    \hspace{5pt}
    \begin{subfigure}[t]{0.25\textwidth}
        \centering
        \includegraphics[width=3.7cm, trim=1.2cm 0 0 0 ,clip]{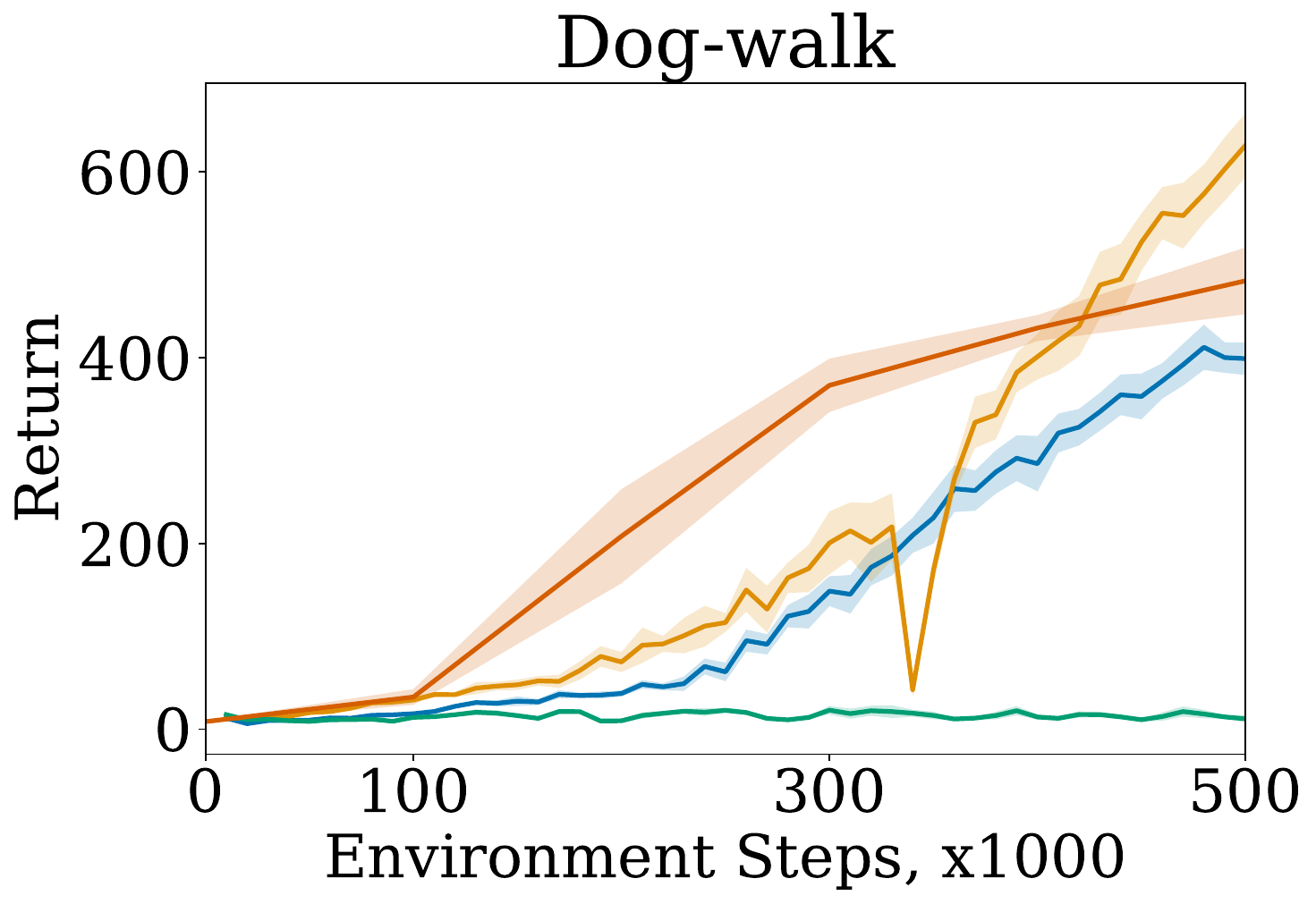}
        \label{subfig:dog_walk}
    \end{subfigure}%
    \hspace{-5pt}
    \begin{subfigure}[t]{0.25\textwidth}
    \centering
        \includegraphics[width=3.7cm, trim=1.2cm 0 0 0 ,clip]{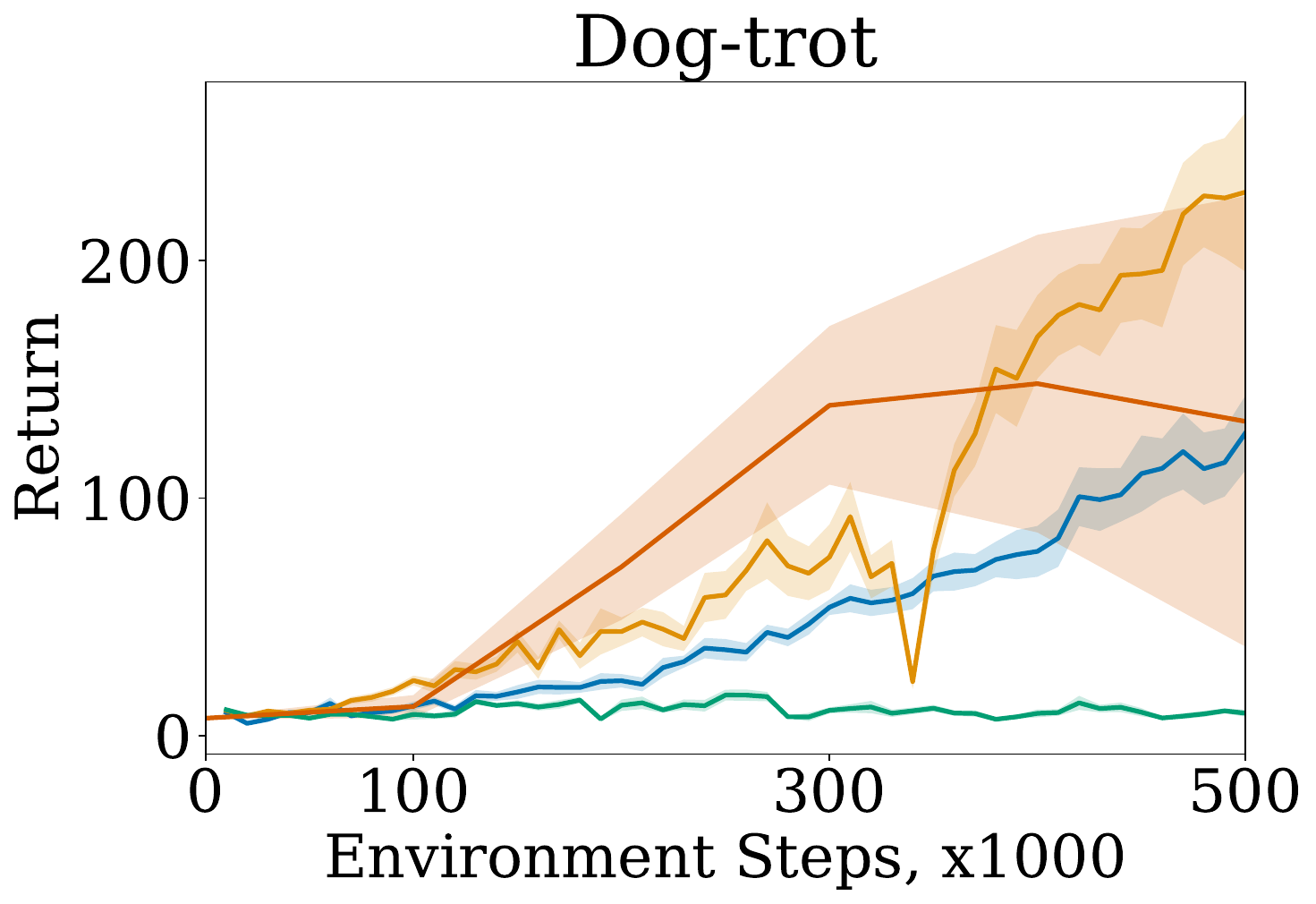}
        \label{subfig:dog_trot}
    \end{subfigure}%
    \hspace{-5pt}
    \begin{subfigure}[t]{0.25\textwidth}
        \centering
        \includegraphics[width=3.7cm, trim=1.2cm 0 0 0 ,clip]{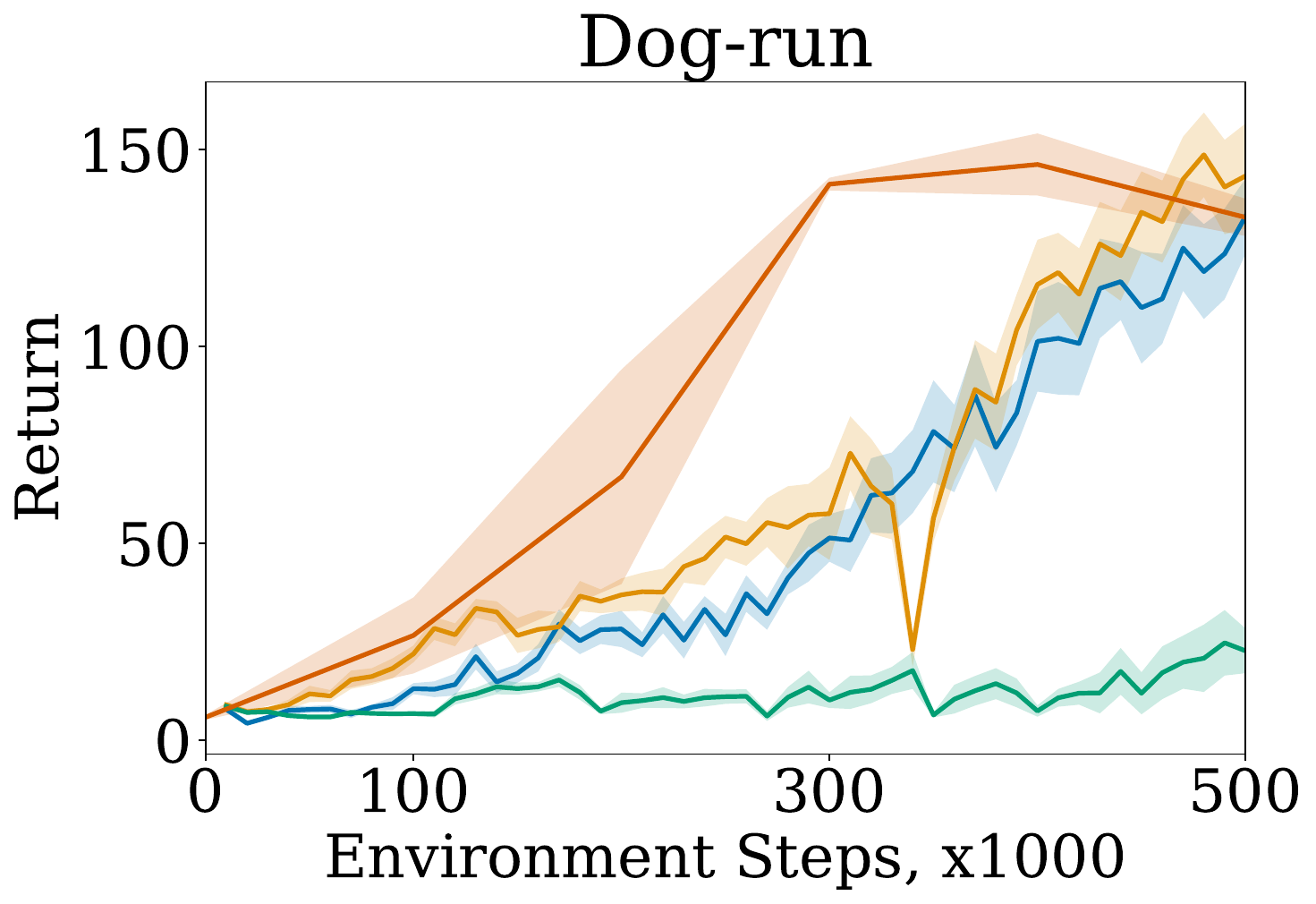}
        \label{subfig:dog_run}
    \end{subfigure}%
    \hspace{-20pt}
    \caption{Mean return on the dog DMC tasks, comparing OFN to SAC with resets and the model-based TD-MPC2. Shaded regions indicate standard error. OFN outperforms SAC with resets, which is unable to learn and OFN with $\mathrm{UTD}=8$ and resetting is competitive with TD-MPC2.}
    \label{fig:all_dog}
\end{figure}

To answer \textbf{Q3}, we move to a set of training environments that is considered exceptionally hard for model-free approaches, namely the dog tasks of the DMC suite. Standard SAC can generally not obtain any reasonable reward and, due to their complex dynamics, these tasks are often tackled using model-based approaches such as TD-MPC2~\citep{hansen2024tdmpc} with complicated update procedures and carefully tuned network architectures. We evaluate SAC and OFN on the dog tasks and compare against TD-MPC2 in Figure~\ref{fig:all_dog}.

First, observe that resetting without OFN obtains no improvement over a random policy. However, OFN with $\mathrm{UTD}=1$ can already obtain very good performance across all tasks, indicating that a major problem for SAC in these high-dimensional tasks is value divergence. When increasing the update ratio to $8$ and adding resetting, we improve the performance of the OFN agent even further and can match the reported results of the strong model-based TD-MPC2 baseline. 

We have already seen that resetting can take care of multiple optimization failures. However, these experiments also indicate that resetting is not a panacea as it is only effective when the initially learned policy can obtain some reward before being overwritten. 
This seems intuitive since resetting to a policy that cannot gather any useful data should not help. 
These results highlight that the early training dynamics of RL are highly important when it comes to training on complex environments and fitting early data correctly and quickly is crucial for success. 

This also opens up the question why resetting in the humanoid environments in the previous sections can yield success even though very little reward is observed. 
Besides greater divergence due to larger observation spaces in the dog MDPs, we suspect that this might be related to the complexity of exploration.
The ability of a random policy to obtain non-trivial reward and information about the environment has been shown to be a crucial factor in explaining the success of DRL methods in discrete environments~\citep{laidlaw2023bridging}, and similar phenomena might be in effect here. 

\section{Related work}

Our work closely examines previous work on the primacy bias and the related resetting technique \citep{anderson1992qlearning,nikishin2022primacy,doro2023barrier,schwarzer2023bigger}.
Going beyond, overestimation and feature learning challenges are a widely studied phenomenon in the literature.

{\bf Combatting overestimation}~~
Overestimation in off-policy value function learning is a well-established problem in the RL literature that dates back far before the prime times of deep learning~\citep{thrun1993issues, precup2001off}. The effects of function approximation error and the effect on variance and bias have been studied~\citep{pendrith1997estimator, mannor2007bias} as well.
With the rise of deep learning, researchers have tried to address the overestimation bias via algorithmic interventions such as combining multiple Q-learning predictors to achieve underestimation~\citep{hasselt2010double, hasselt2016deep, zongzhang2017weighted, lan2020maxmin}, using averages over previous Q-values for variance reduction~\citep{anschel2017averaged}, or general error term correction~\citep{donghun2013bias, fox2016taming}. In the context of actor-critic methods, the twinned critic minimization approach of \citet{fujimoto2018addressing} has become a de-facto standard. Most of these approaches are not applicable or break down under very high update ratios.
To regulate the overestimation-pessimism balance more carefully, several authors have attempted to use larger ensembles of independent Q-value estimates \citep{lee2021sunrise,peer2021ensemble,chen2021randomized,hiraoka2022dropout}. Ensembling ideas were also combined with ideas from distributional RL~\citep{bellemare2017distributional} to combat overestimation~\citep{kuznetsov2020controlling}.  Instead of addressing the statistical bias in deep RL, our study focuses on the problems inherent to neural networks and gradient based optimization for value function estimation. Work from offline-to-online RL has demonstrated that standard layer normalization can bound value estimates during offline training and mitigate extrapolation while still allowing for exploration afterwards~\citep{ball2023efficient}. Layer normalization has subsequently been used to achieve generally strong results in offline RL~\citep{tarasov2023rebrac}. Our work is also related to a recent contribution using batch-normalization for increased computational efficiency by~\citet{bhatt2024crossq} who focus on decreasing update ratios rather than increasing them. A concurrent work by~\citet{nauman2024overestimation} provides a large scale study on different regularization techniques to combat overestimation. This work also demonstrates the efficacy of SAC on the dog tasks when properly regularized but it does not highlight the effects of Q-value divergence from exploding gradients as a key challenge for this set of environments.

{\bf Combating plasticity loss}~~ Another aspect of the primacy bias is the tendency of neural networks to lose their capacity for learning over time~\citep{igl2021transient}, sometimes termed \emph{plasticity loss}~\citep{lyle2021understanding, abbas2023loss}. Recent work mitigates plasticity loss using feature rank maximization~\citep{kumar2021implicit}, regularization~\citep{lyle2023understanding}, or learning a second copied network~\citep{nikishin2024deep}. Some of the loss stems from neurons falling dormant over time~\citep{sokar2023dormant}. A concurrent, closely related work by~\citet{lyle2024disentangling} disentangles the causes for plasticity loss further. They use layer normalization to prevent some of these causes, which is closely related to our unit ball normalization. Our work differs in that we focus on the setting of high update ratios and use stronger constraints to mitigate value divergence rather than plasticity loss. 

\section{Conclusion and future work}

By dissecting the effects underlying the primacy bias, we have identified a crucial challenge: \emph{value divergence}. 
While the main focus in studying increased Q-value has been on the statistical bias inherent in off-policy sampling, we show that Q-value divergence can arise due to problems inherent to neural network optimization.
This optimization-caused divergence can be mitigated using the unit-ball normalization approach, which shines on the \textsf{dm\_control} benchmark with its simplicity and efficacy. 
With this result, we challenge the assumption that failure to learn in high-UTD settings primarily stems from \emph{overfitting} early data by showing that combating value divergence is competitive with resetting networks. 
This offers a starting point towards explaining the challenges of high-UTD training in more detail and opens the path towards even more performant and sample efficient RL in the future.

However, as our other experiments show, mitigating value overestimation through optimization is not the only problem that plagues high-UTD learning. 
To clearly highlight these possible directions for future work, we provide an extensive discussion of open problems in Appendix~\ref{app:open}. 
Additional problems, such as \emph{exploration failures} or \emph{suboptimal feature learning}, can still exist and need to be resolved to unlock the full potential of high-UTD RL.

\clearpage

\section*{Acknowledgements}

EE and MH's research was partially supported by the Army Research Office under MURI award W911NF20-1-0080, the DARPA Triage Challenge under award HR001123S0011, and by the University of Pennsylvania ASSET center. Any opinions, findings, and conclusion or recommendations expressed in this material are those of the authors and do not necessarily reflect the view of DARPA, the Army, or the US government.
AMF acknowledges the funding from the Canada CIFAR AI Chairs program, as well as the support of the Natural Sciences and Engineering Research Council of Canada (NSERC) through the Discovery Grant program (2021-03701).
Resources used by CV, AMF, and IG in preparing this research were provided, in part, by the Province of Ontario, the Government of Canada through CIFAR, and companies sponsoring the Vector Institute.

The authors thank the members of the GRASP lab at UPenn, and the members of the AdAge and TISL labs at UofT, as well as the anonymous reviewers for their valuable feedback.

\bibliographystyle{template/rlc}
\bibliography{references}

\newpage

\appendix

\section{Implementation details and hyperparameters} \label{app:impl}

We employ two commonly used implementations, one for fast iterations on priming experiments (\href{https://github.com/denisyarats/pytorch\_sac}{https://github.com/denisyarats/pytorch\_sac}) and one for scaling up our experiments to high update ratios (\href{https://github.com/proceduralia/high\_replay\_ratio\_continuous\_control}{https://github.com/proceduralia/high\_replay\_ratio\_continuous\_control}). All experiments in the main sections use default hyperparameters of the high update ratio codebase  unless otherwise specified with minor exceptions.

\begin{table}[H]
    \parbox[t]{.45\linewidth}{
    \label{tab:sharedhparam}
    \centering
    \caption{Shared hyperparameters between priming and high-UTD implementations}
    \begin{tabular}{ | m{3.5cm} | m{2cm}| }
      \hline
      Optimizer & Adam \\ 
      \hline
      Adam $\beta_1$ & $0.9$ \\ 
      \hline
      Adam $\beta_2$ & $0.999$ \\ 
      \hline
      Adam $\varepsilon$ & $1e-8$ \\ 
      \hline
      Actor Learning Rate & $4e-3$ \\ 
      \hline
      Critic Learning Rate & $4e-3$ \\ 
      \hline
      Temp. Learning Rate & $3e-4$ \\ 
      \hline
      Batch Size
      & $256$ \\ 
      \hline
      $\gamma$ & $0.99$ \\
      \hline
      $\tau$ & $0.005$ \\
      \hline
      \# critics & $2$ \\
      \hline
      \# critic layers & $2$ \\
      \hline
      \# actor layers & $2$ \\
     \hline
      critic hidden dim & $256$ \\
     \hline
      actor hidden dim & $256$ \\
     \hline
    \end{tabular}
    \label{tab:shared}
    }
    \hfill
    \parbox[t]{.55\linewidth}{
    \centering
    \caption{Differing hyperparameters between priming and high-UTD implementations}
    \begin{tabular}{ |m{2cm} | m{2.5cm} |  m{2.5cm}| }
     \hline
     & Priming & High UTD \\
     \hline\hline
     Initial \newline temperature & $0.1$ & $1.0$ \\
     \hline
     Target \newline entropy & -action\_dim & -action\_dim / 2 \\
     \hline
     actor log \newline std bounds & [-5, 2] & [-10, 2] \\
     \hline
    \end{tabular}
    }
\end{table}

\section{Additional priming experiments} \label{app:priming}

\subsection{Activation functions}

\begin{figure}[H]
\centering
    \begin{subfigure}[b]{0.8\textwidth}
        \centering
        \includegraphics[height=0.4cm]{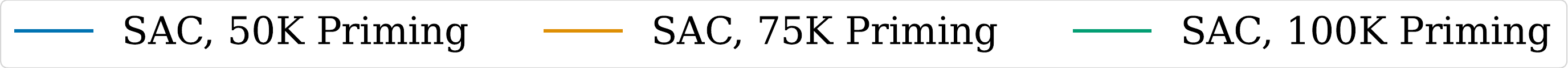}
    \end{subfigure}\\%
    \begin{subfigure}[b]{0.33\textwidth}
        \centering
        \includegraphics[width=4.7cm, trim=1cm 1cm 1cm 1cm ,clip]{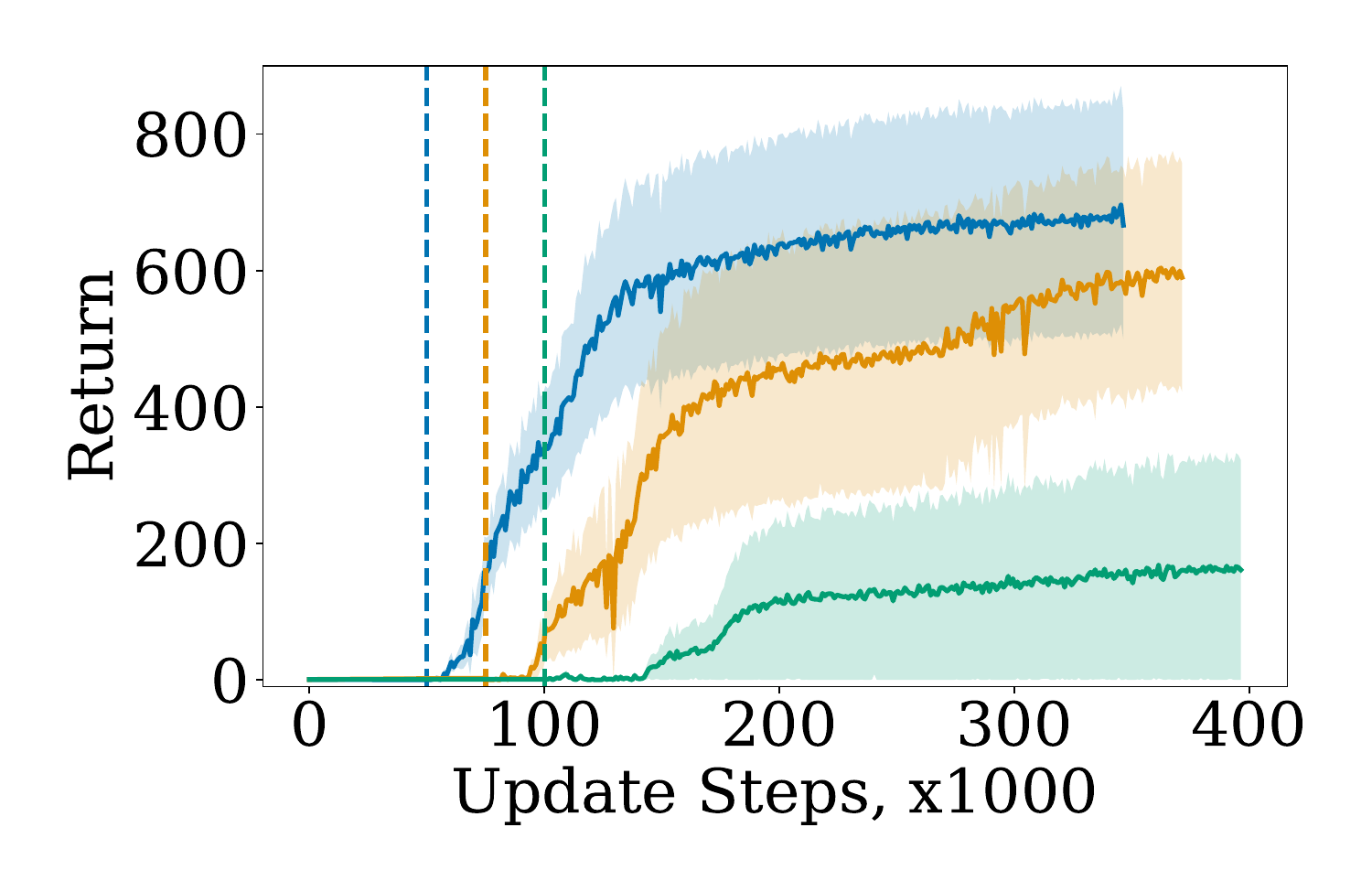}
        \label{subfig:elu_priming_base_ret}
    \end{subfigure}%
    \begin{subfigure}[b]{0.33\textwidth}
    \centering
        \includegraphics[width=4.7cm, trim=1cm 1cm 1cm 1cm ,clip]{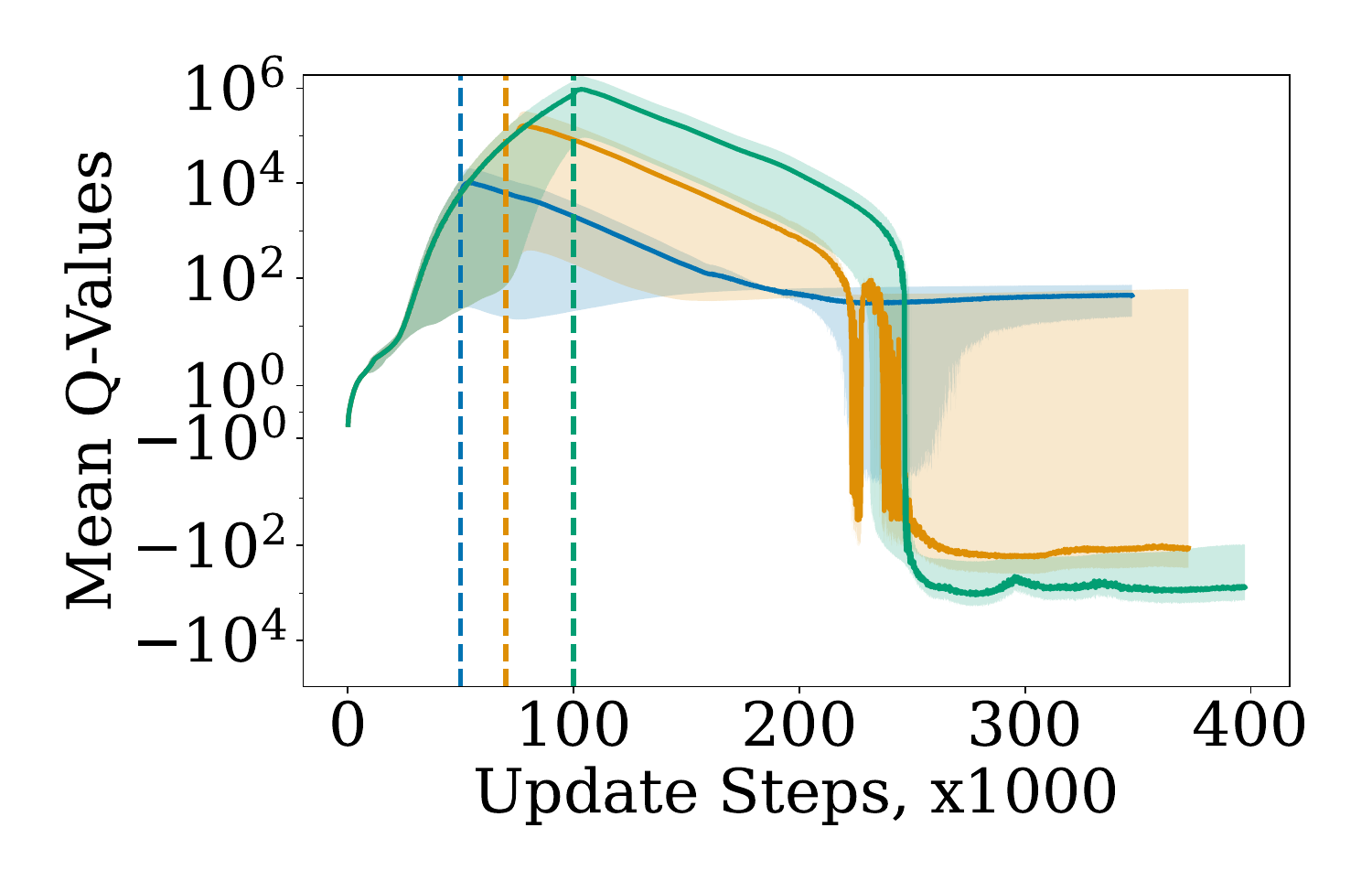}
        \label{subfig:priming_base_Q}
    \end{subfigure}%
    \begin{subfigure}[b]{0.33\textwidth}
        \centering
        \includegraphics[width=4.7cm, trim=1cm 1cm 1cm 1cm ,clip]{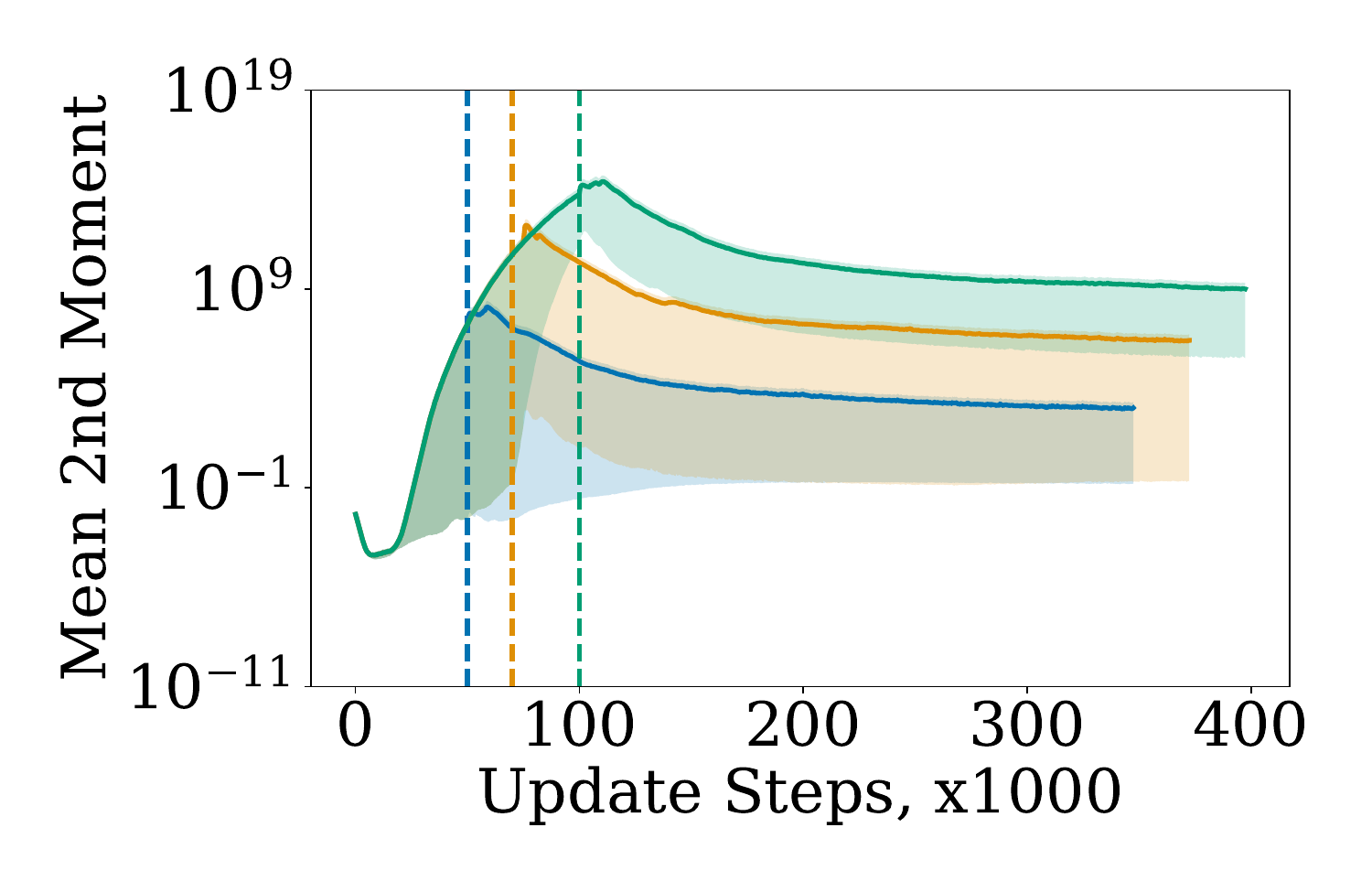}
        \label{subfig:elu_priming_base_mom}
    \end{subfigure}%
    \vspace{-5pt}
    \caption{ELU activations. Return, in-distribution Q-values and Adam optimizer moments during priming for different lengths. Dotted lines correspond to end of priming. More priming leads to lower return and larger Q-value and optimizer divergence.}
    \label{fig:elu_priming_base}
\end{figure}

During our experiments, we found that the ReLU activation can sometimes lead to destabilization of other parts of the SAC agent during priming. We found that using ELU~\citep{clevert2016accurate} activations instead remedies some of these issues. We repeat various experiments from Section~\ref{sec:investigating} again but with more stable activations. First, we show in Figure~\ref{fig:elu_priming_base} that divergence happens similar to before and that it is correlated with the amount of priming.

\begin{figure}[H]
\begin{minipage}[b]{.35\textwidth}
\centering
    \begin{subfigure}[b]{\textwidth}
        \centering
        \includegraphics[height=0.8cm]{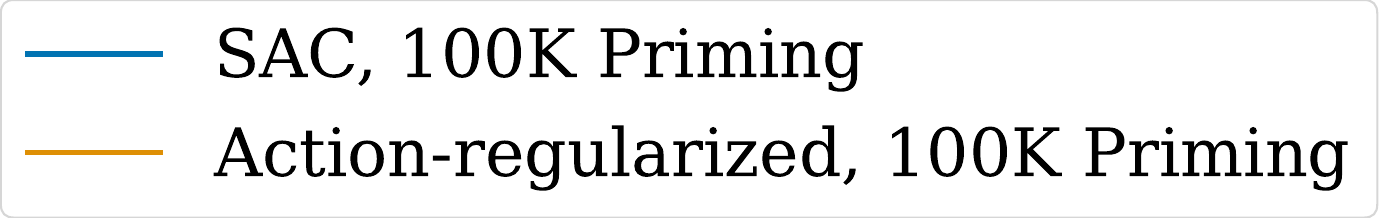}
    \end{subfigure}\\%
    \begin{subfigure}[b]{\textwidth}
        \centering
        \includegraphics[width=4.8cm, trim=1cm 1cm 1cm 1cm ,clip]{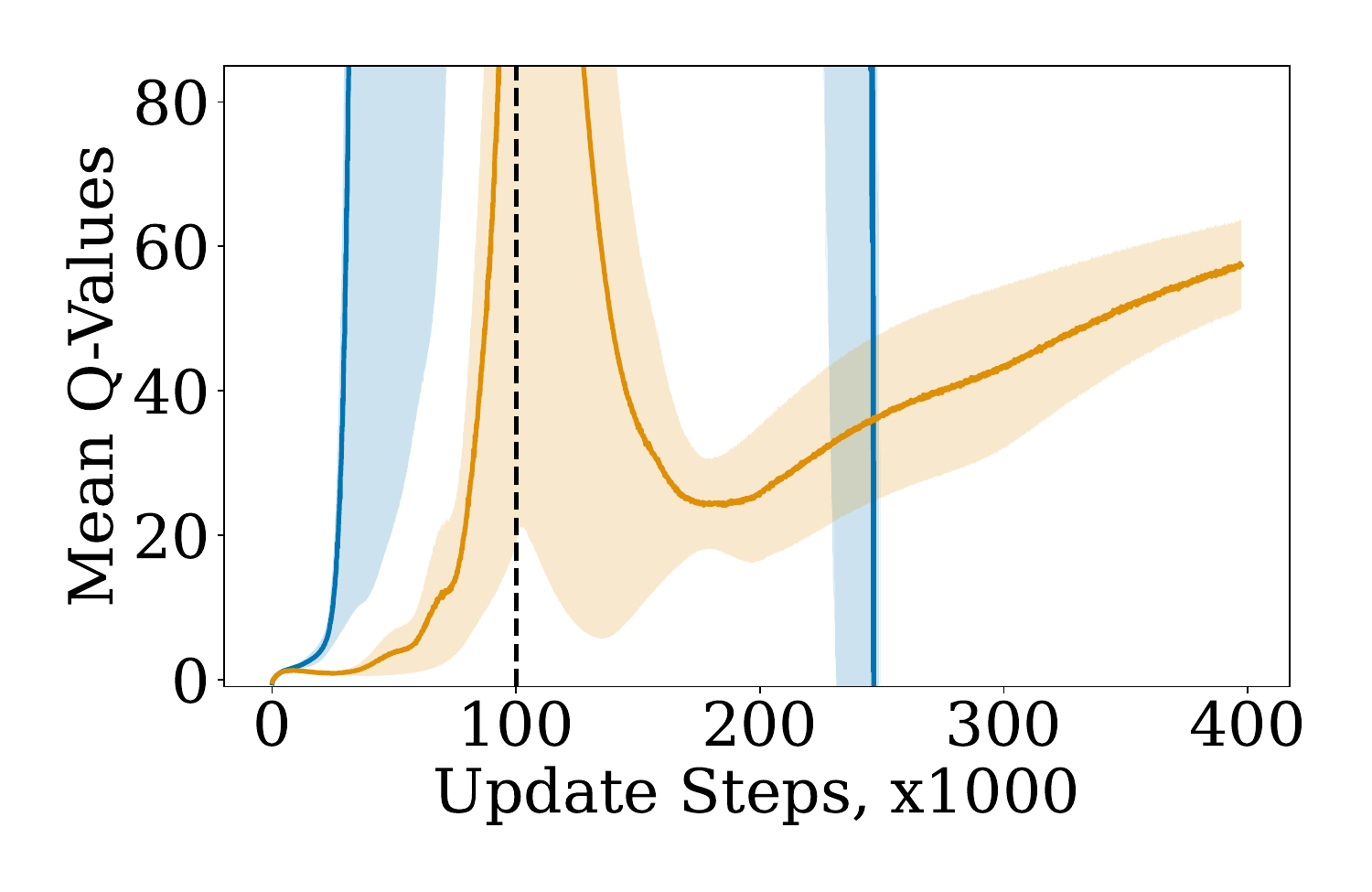}
        \label{subfig:elu_priming_causes_Q}
    \end{subfigure}%
    \vspace{-5pt}
    \caption{ELU activations. Priming with SAC and action regularization during priming. The latter lowers divergence. }
    \label{fig:elu_priming_causes}
\end{minipage}
\hfill
\begin{minipage}[b]{.62\textwidth}
    \centering
    \begin{subfigure}[b]{0.8\textwidth}
        \centering
        \includegraphics[height=0.4cm]{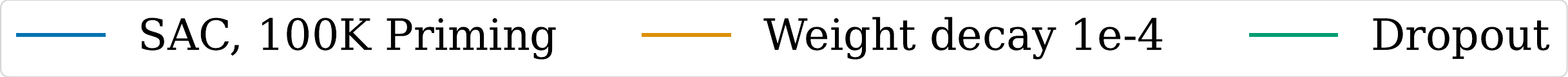}
    \end{subfigure}\\%
    \begin{subfigure}[b]{0.5\textwidth}
        \centering
        \includegraphics[width=4.8cm, trim=1cm 1cm 1cm 1cm ,clip]{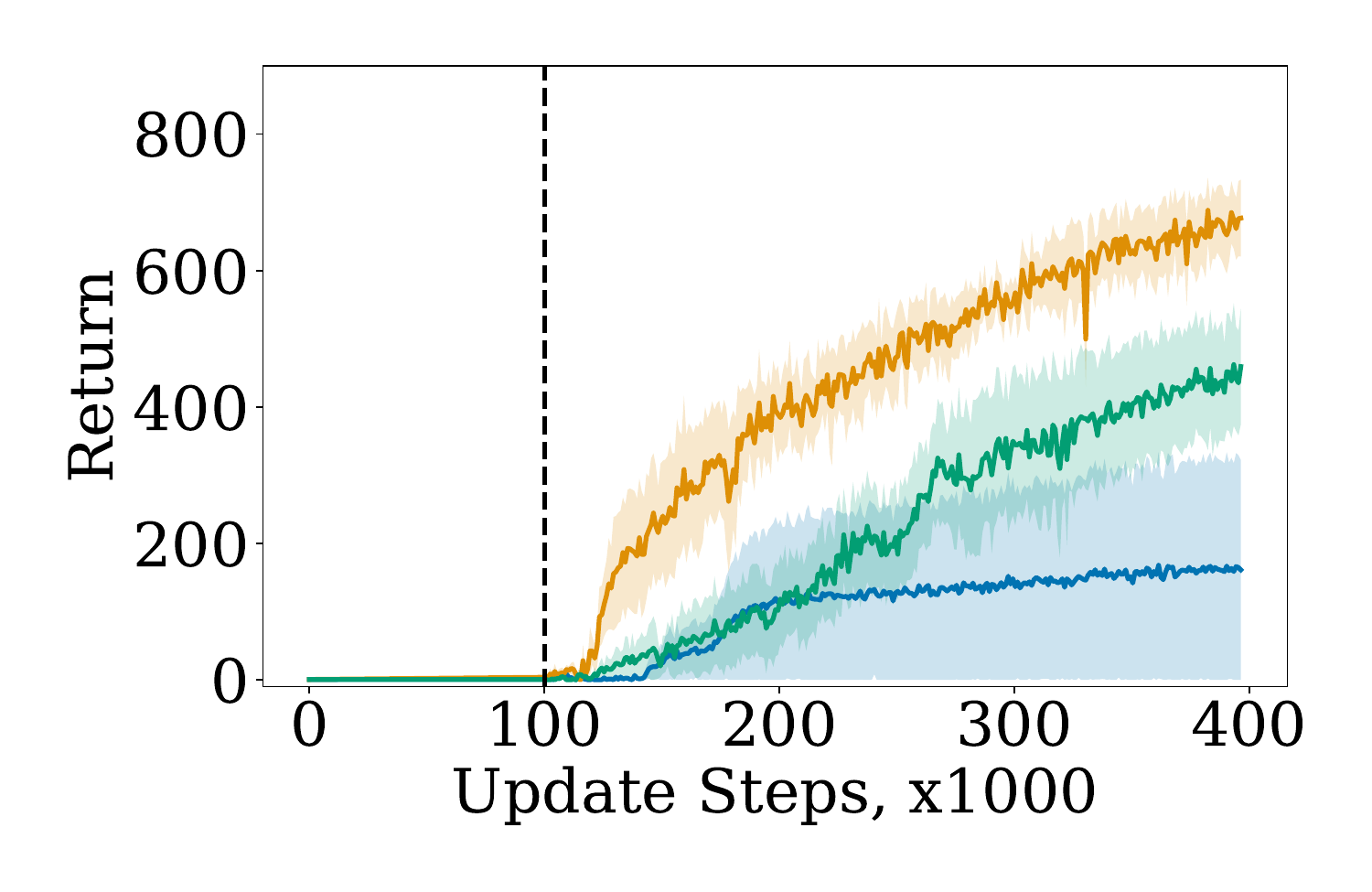}
        \label{subfig:elu_priming_abl_ret}
    \end{subfigure}%
    \begin{subfigure}[b]{0.5\textwidth}
    \centering
        \includegraphics[width=4.8cm, trim=1cm 1cm 1cm 1cm ,clip]{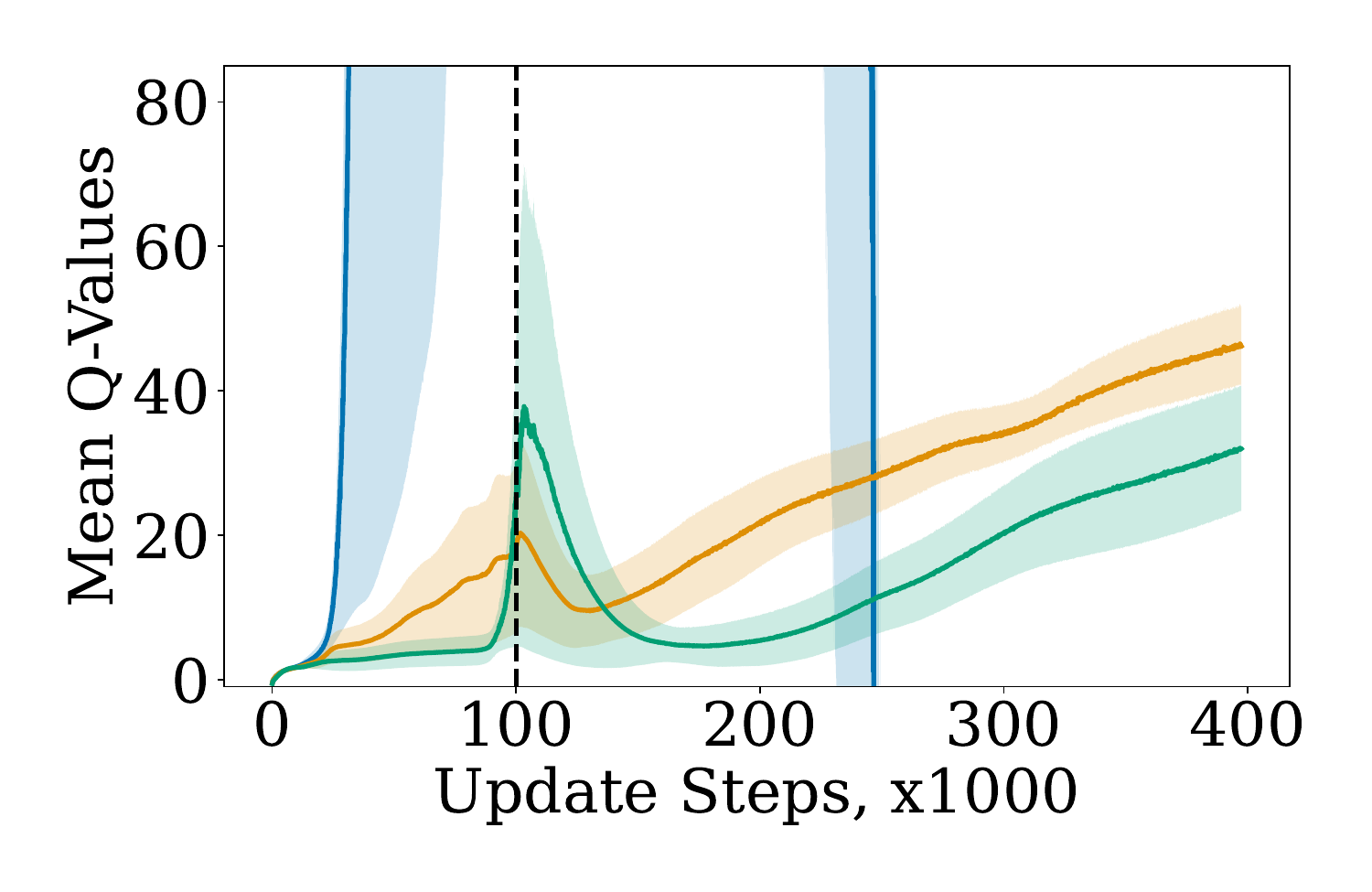}
        \label{subfig:elu_priming_abl_Q}
    \end{subfigure}%
    \vspace{-18pt}
    \caption{ELU activations. Return and
    Q-values of priming runs with weight decay and dropout. Results indicate that both regularizations mitigate priming more than with ReLUs. }
    \label{fig:elu_priming_abl}
\end{minipage}
    \vspace{-5pt}
\end{figure}

Furthermore, we discussed that the divergence is most likely triggered by out of distribution action prediction (see Figure~\ref{fig:elu_priming_causes}) and that regularization can help. When using ELUs, the effect of regularization is much more stable and as expected but still not as good as our OFN approach from Section~\ref{sec:evalmethod} (compare with Figure~\ref{fig:elu_priming_abl}). Dropout leads to significantly worse performance and L$^2$ regularization learns Q-values too small for the obtained return which we suspect correlates with decreased exploration.

\subsection{Optimizer divergence} \label{app:priming_opt}

With more stable effects from the ELU activation, we introduce a second intervention to the priming stage. We hypothesize that most of the divergence stems from the second optimizer term that will propell the gradients to increase more and more over time. To test this, we run an additional experiment in which we use standard stochastic gradient descent (SGD)~\citep{robbins1951stochastic} with first-order momentum~\citep{rumelhart1986learning, sutskever2013on} during priming to isolate the effect of the second-order momentum term. We compare this against RMSProp which is equivalent to Adam but without the first optimizer term instead. The results are shown in Figure~\ref{app:priming_opt}. As we can see, the divergence is almost completely gone when using SGD with momentum but is even larger in RMSProp. Note that running the same experiment with ReLU activations  leads to divergence in the actor when using SGD only. We suspect this might have to do with divergence in the actor entropy.

\begin{figure}[H]
\centering
    \begin{subfigure}[b]{0.8\textwidth}
        \centering
        \includegraphics[height=0.8cm]{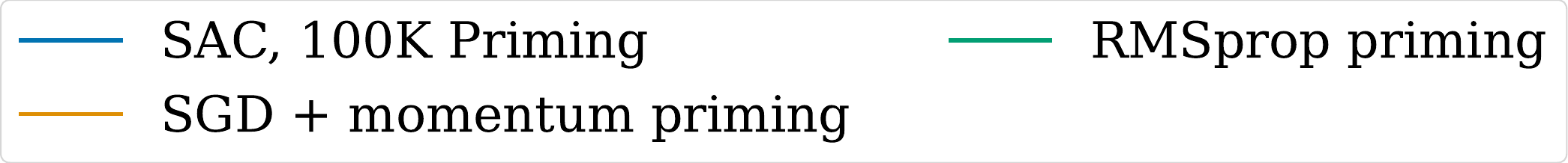}
    \end{subfigure}\\%
    \begin{subfigure}[b]{0.33\textwidth}
        \centering
        \includegraphics[width=4.7cm, trim=1cm 1cm 1cm 1cm ,clip]{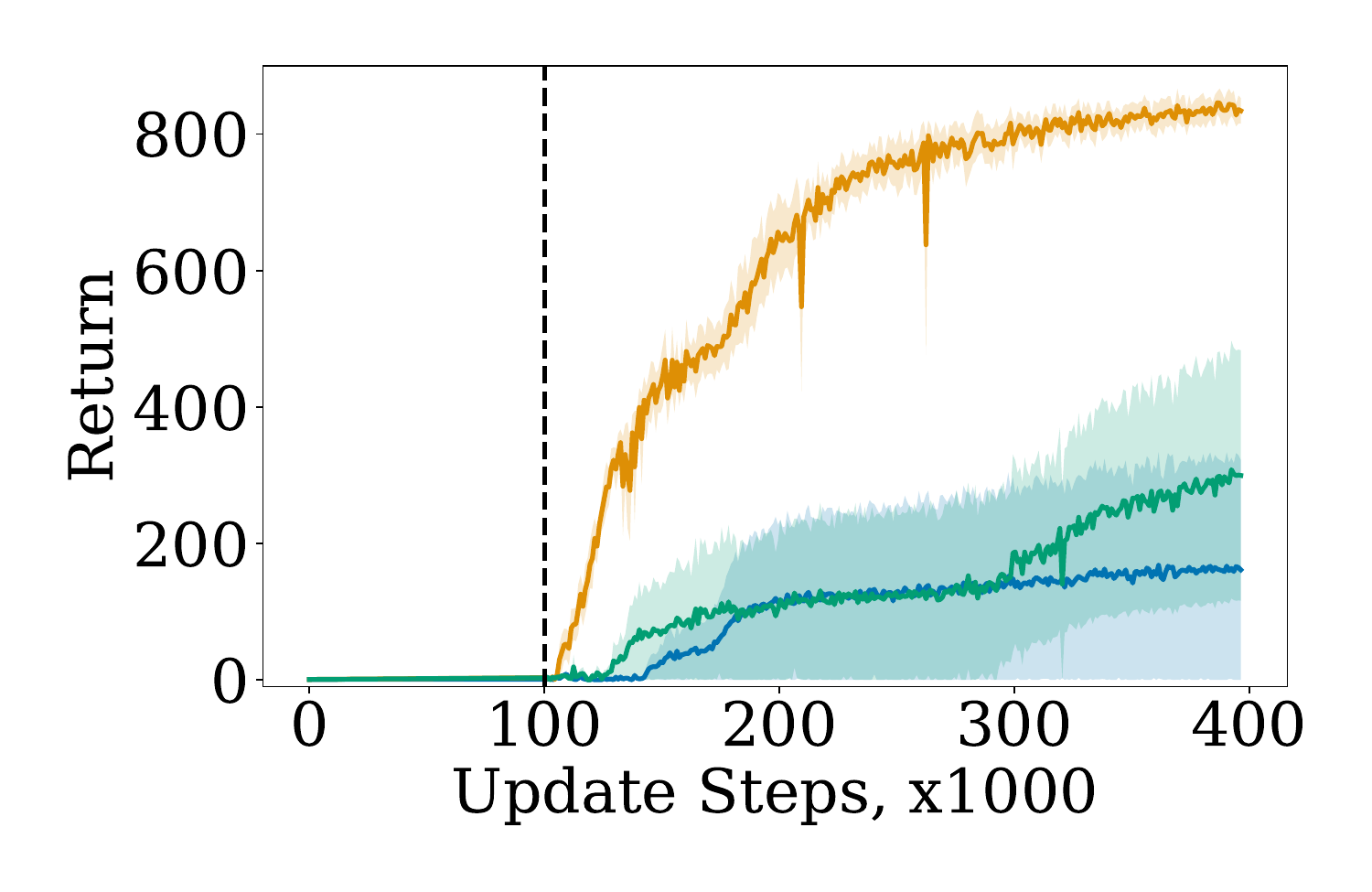}
        \label{subfig:priming_opt_ret}
    \end{subfigure}%
    \begin{subfigure}[b]{0.33\textwidth}
    \centering
        \includegraphics[width=4.7cm, trim=1cm 1cm 1cm 1cm ,clip]{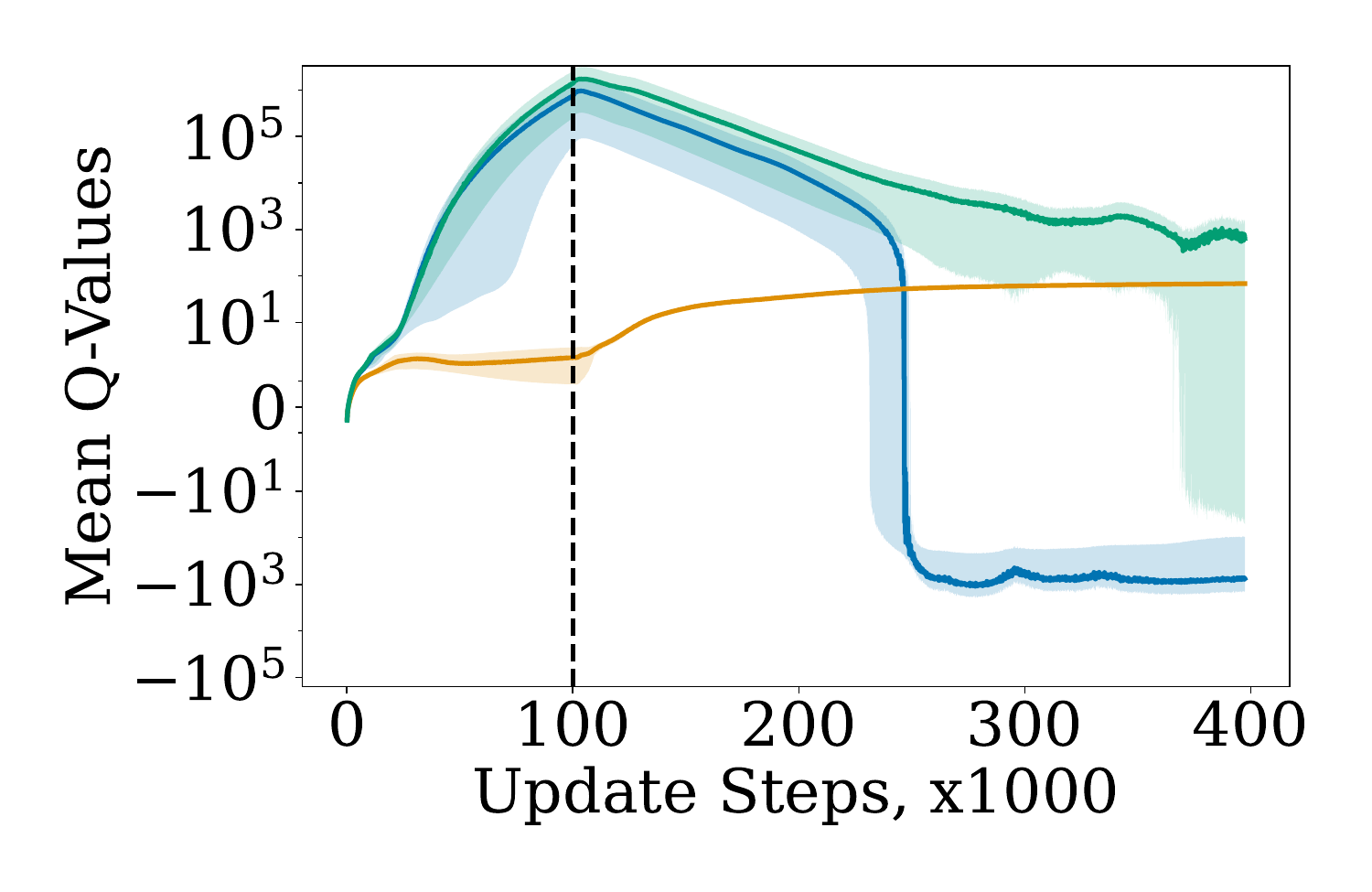}
        \label{subfig:priming_opt_Q}
    \end{subfigure}%
    \vspace{-5pt}
    \caption{Comparing standard SAC priming to priming when using either SGD+momentum or RMSProp during the priming updates. SGD+momentum does not diverge with ELU activations, indicating that the second-order momentum term is the problematic one.}
    \label{fig:priming_opt}
\end{figure}

\subsection{Effective dimension} \label{app:priming_dim}

Let $\Phi \in \mathbb{R}^{|\mathcal{S}| |\mathcal{A}| \times d}$ be a feature matrix (in our case produced by $\phi$). The effective dimension of a feature matrix has previously been defined as 
\begin{equation*}
    \text{srank}_{\delta} = \min 
    \left\{ k : \cfrac{\sum_{i=1}^k \sigma_i(\Phi)}{\sum_{i=1}^d \sigma_i(\Phi) \geq 1 - \delta}
    \right\} \enspace ,
\end{equation*}
where $\delta$ is a threshold parameters and $\{\sigma_i(\Phi)\}$ are the singular values of $\Phi$ in decreasing order~\citep{yang2020harnessing, kumar2021implicit}.

\begin{figure}[H]
\centering
    \begin{subfigure}[b]{0.8\textwidth}
        \centering
        \includegraphics[height=0.4cm]{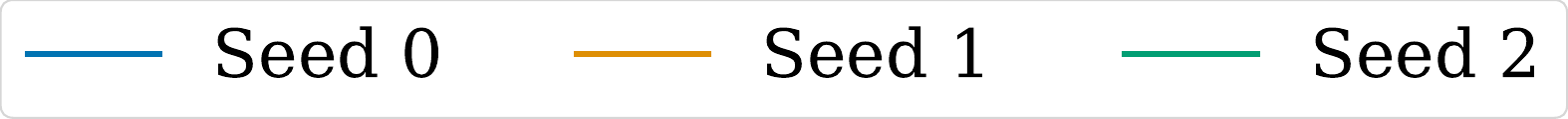}
    \end{subfigure}\\%
    \begin{subfigure}[b]{0.33\textwidth}
        \centering
        \includegraphics[width=4.7cm, trim=1cm 1cm 1cm 1cm ,clip]{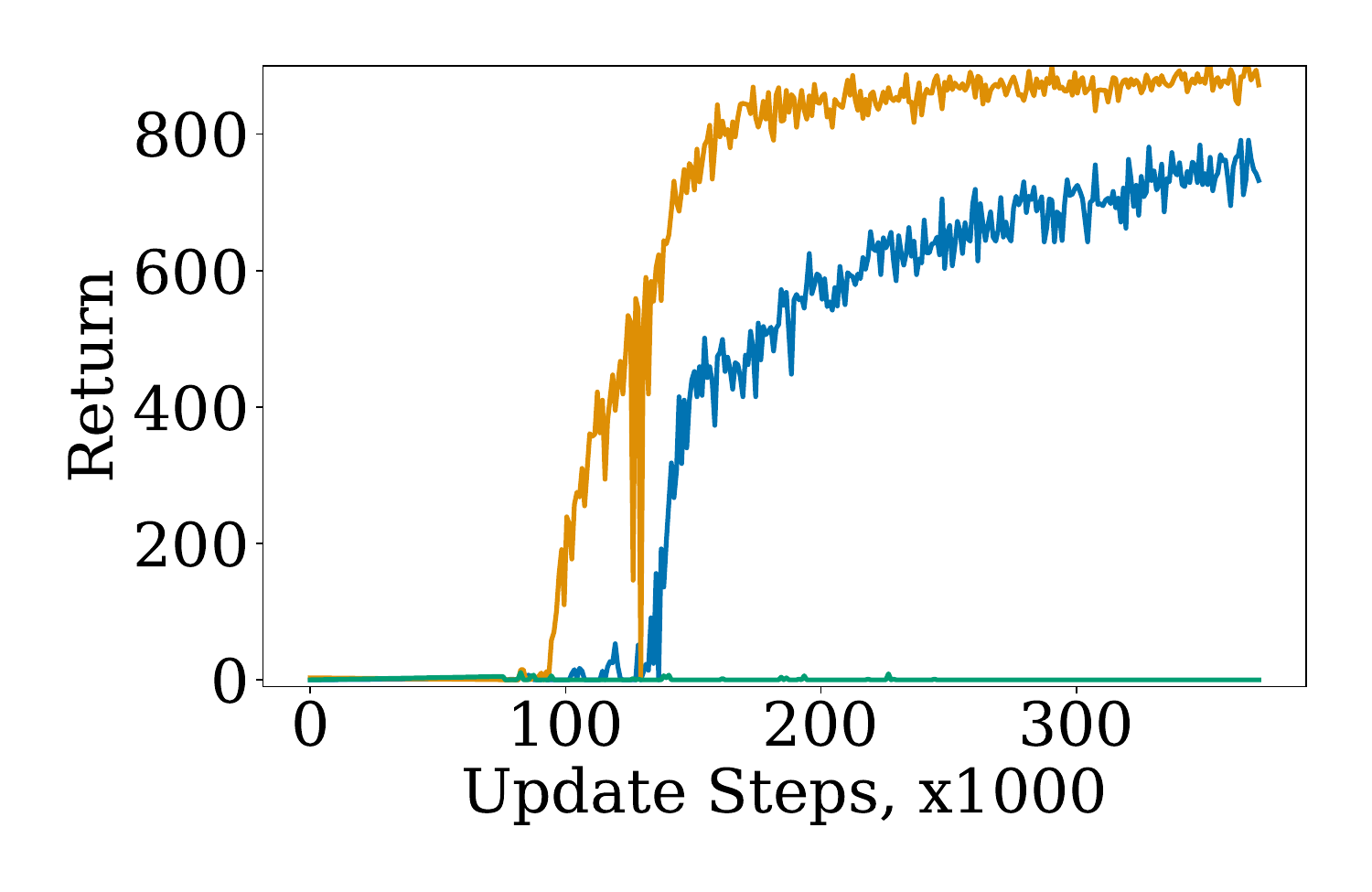}
        \label{subfig:elu_dim_ret}
    \end{subfigure}%
    \begin{subfigure}[b]{0.33\textwidth}
    \centering
        \includegraphics[width=4.7cm, trim=1cm 1cm 1cm 1cm ,clip]{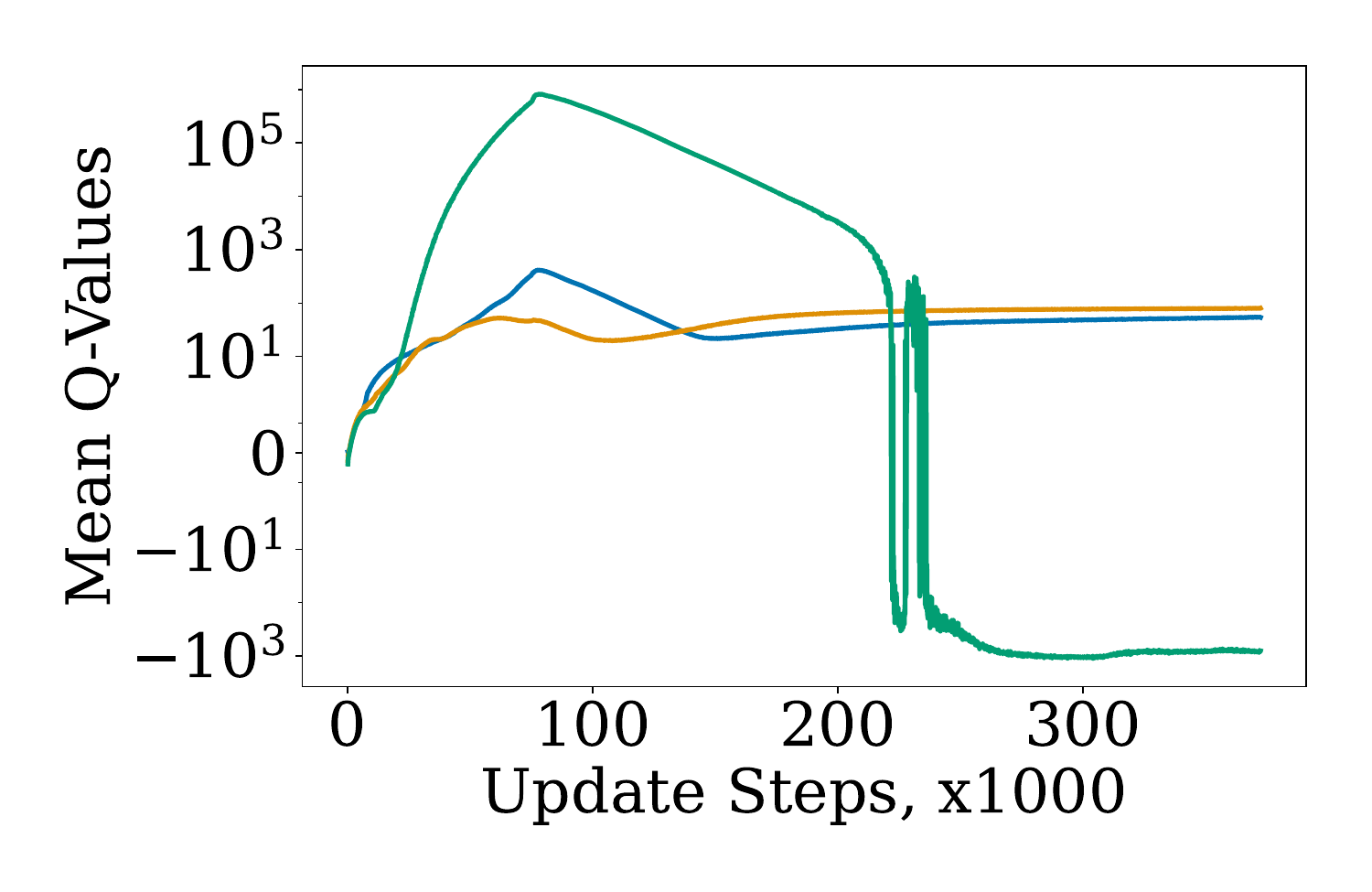}
        \label{subfig:priming_dim_Q}
    \end{subfigure}%
    \begin{subfigure}[b]{0.33\textwidth}
        \centering
        \includegraphics[width=4.7cm, trim=1cm 1cm 1cm 1cm ,clip]{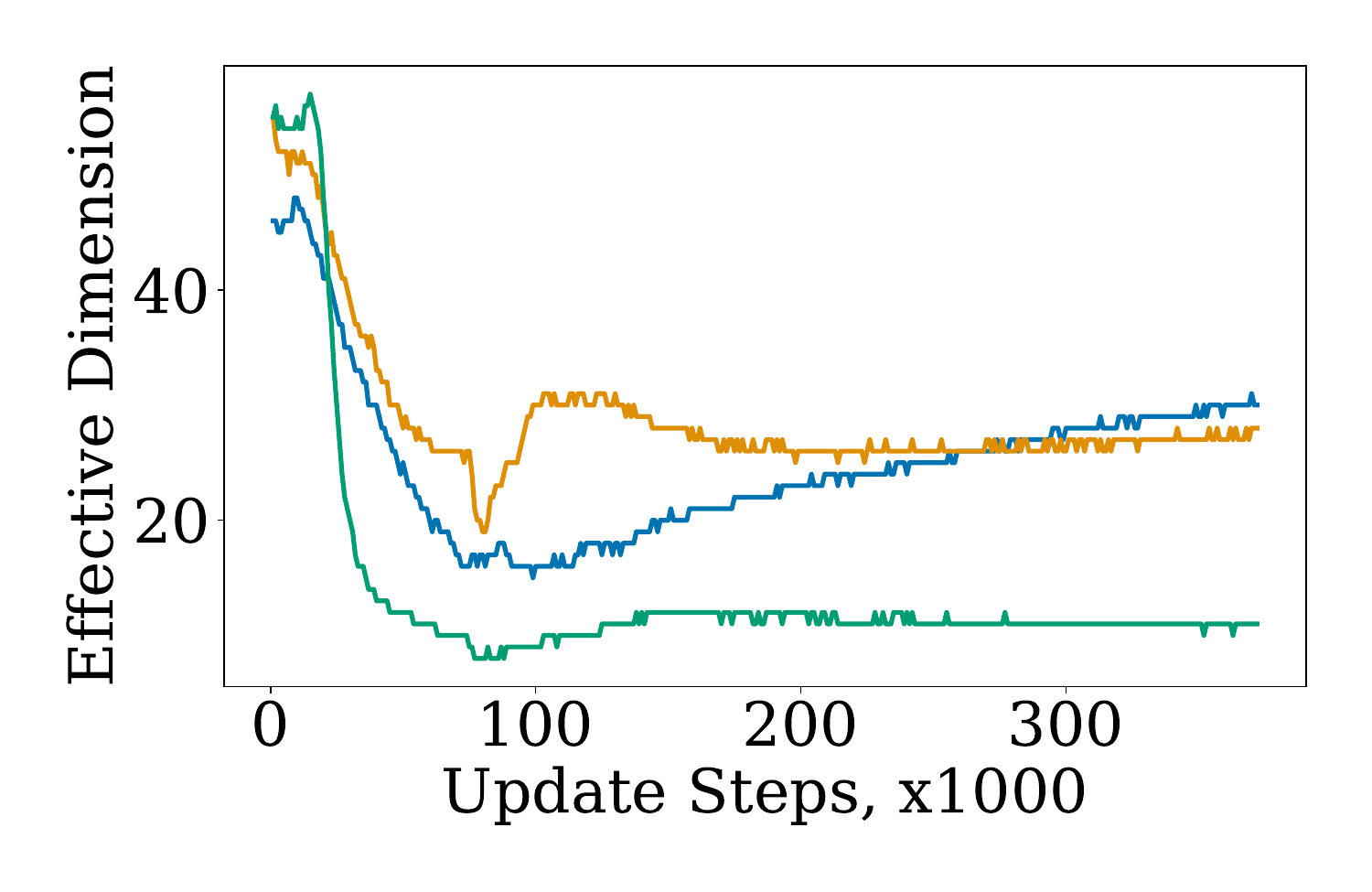}
        \label{subfig:elu_priming_dim_ed}
    \end{subfigure}%
    \vspace{-5pt}
    \caption{Returns, Mean Q-values and effective dimension for $3$ seeds of standard priming for 75,000 steps. When divergence happens, effective dimension is lost. If the effective dimension drops too far, the agent has difficulties to recover.}
    \label{fig:elu_priming_dim}
\end{figure}

An additional finding of ours is that divergence of Q-values is correlated with this effective rank $\text{srank}_{\delta}$. We plot three different random seeds that have been subjected to 75,000 steps of priming in Figure~\ref{fig:elu_priming_dim}; the effective rank is approximated over a sample 10 times the size of the embedding dimension. We observe, that divergence correlates with a decrease in effective dimension and that when divergence is exceptionally strong, the effective dimension drops so low that the agent has trouble to continue learning. This might explain the failure to learn observed by~\citet{nikishin2022primacy}. However, as long as the effective dimension does not drop too far, the agent can recover and regain capacity by observing new data. 
Previous work on effective rank loss has often assumed that it is mostly irreversible, yet we find that this is not always the case.
We suspect that in complete failure cases, the policy has collapse and rarely any new data is seen.

\newpage

\section{Additional experimental results} \label{app:exp}

\subsection{Returns on all environments} \label{app:exp_ret}

\begin{figure}[H]
\centering
    \begin{subfigure}[b]{0.8\textwidth}
        \centering
        \includegraphics[height=0.8cm]{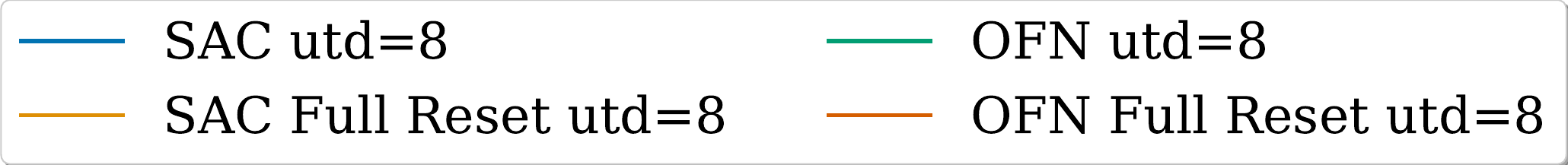}
    \end{subfigure}\\%
    \begin{subfigure}[b]{1\textwidth}
        \centering
        \includegraphics[width=15cm, trim=0cm 0cm 0cm 0cm ,clip]{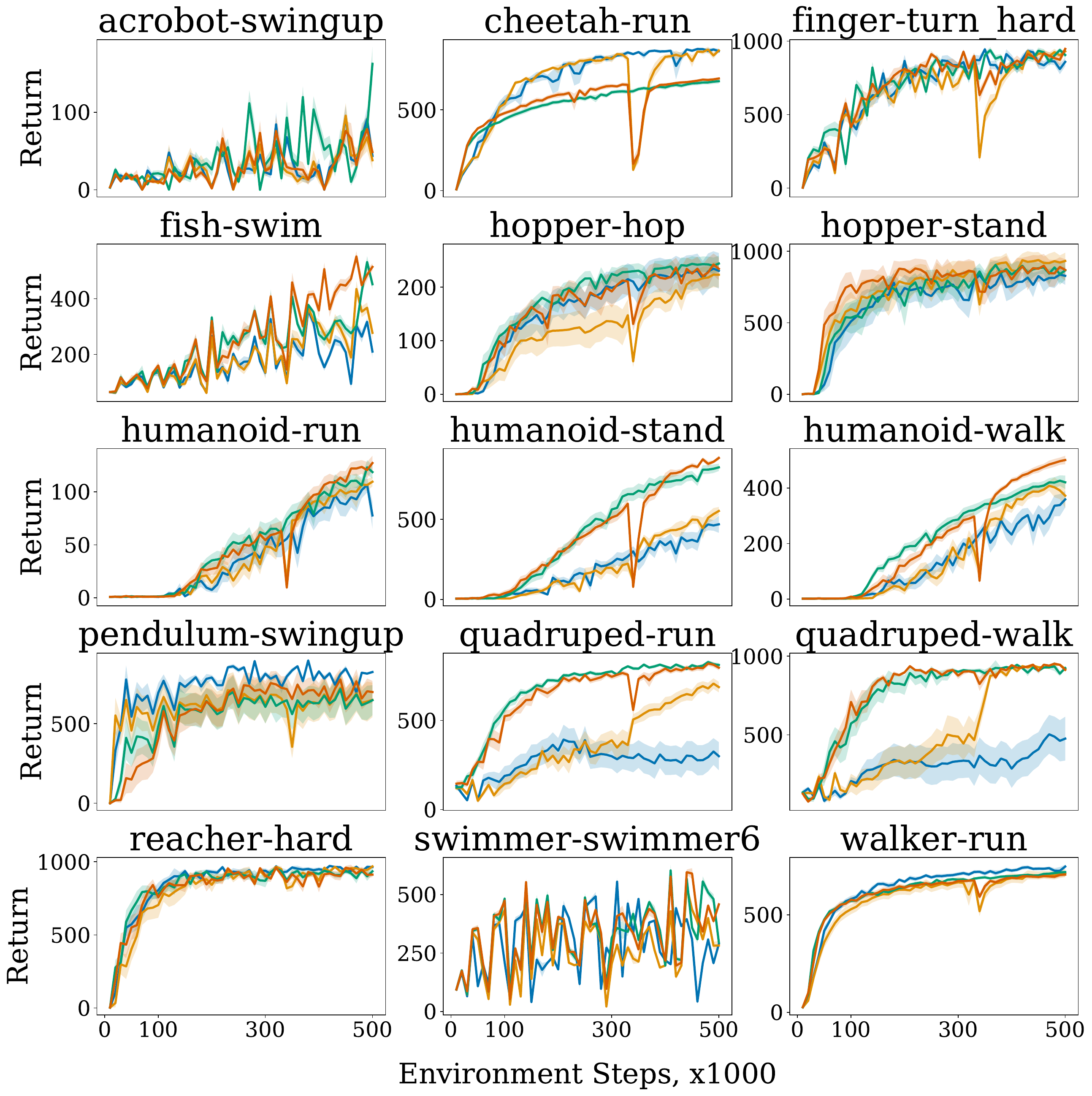}
    \end{subfigure}%
    \vspace{-5pt}
    \caption{UTD8 Returns on Full DMC15-500K.}
    \label{fig:utd8_ret}
\end{figure}

\newpage

\begin{figure}[H]
\centering
    \begin{subfigure}[b]{0.8\textwidth}
        \centering
        \includegraphics[height=0.8cm]{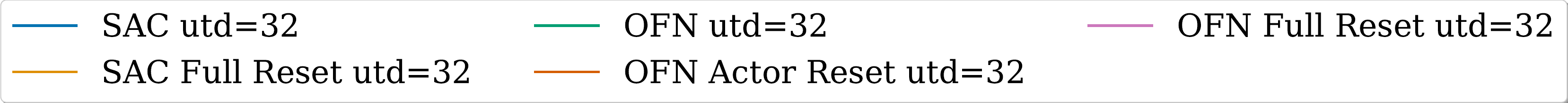}
    \end{subfigure}\\%
    \begin{subfigure}[b]{1\textwidth}
        \centering
        \includegraphics[width=15cm, trim=0cm 0cm 0cm 0cm ,clip]{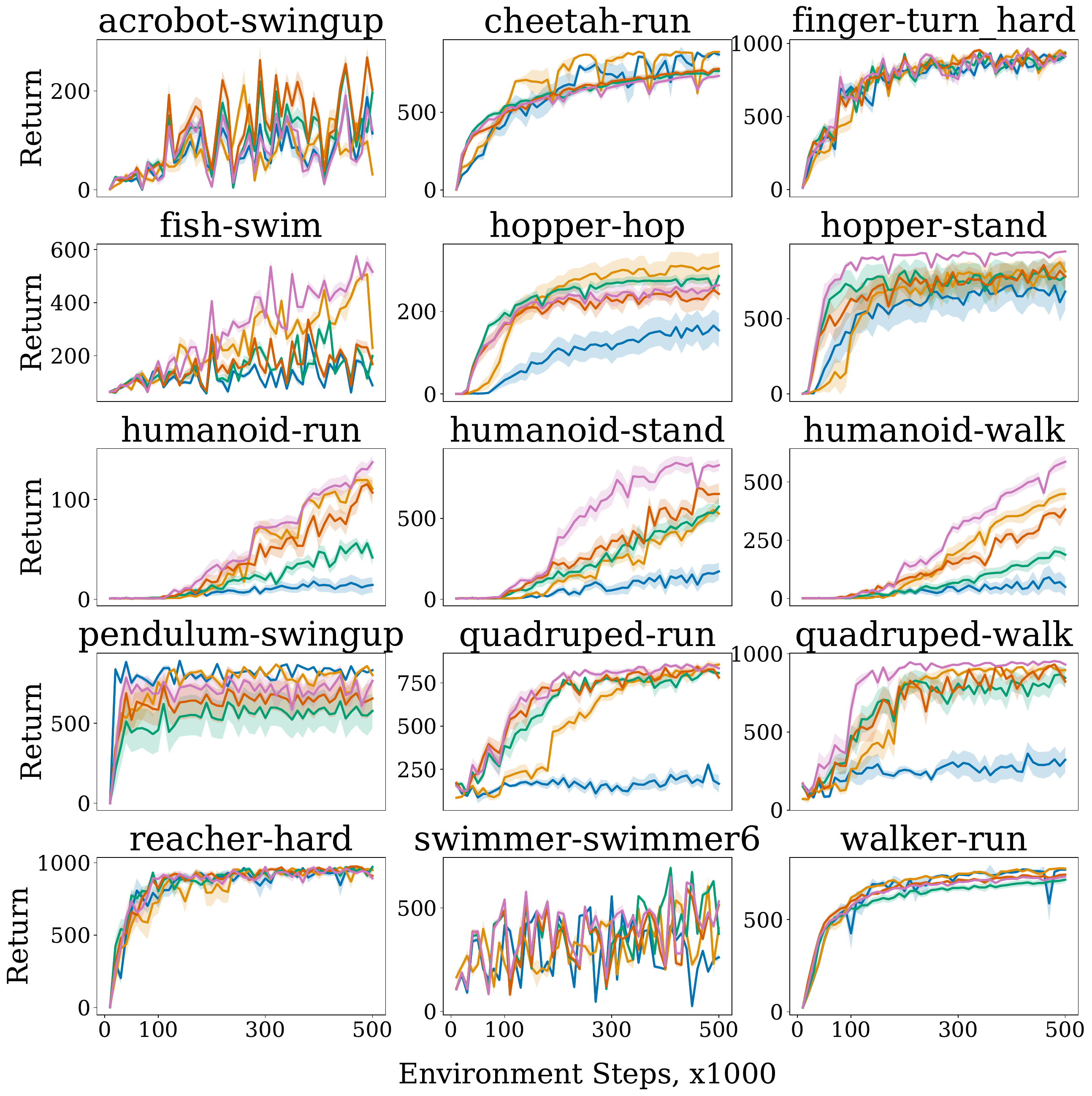}
    \end{subfigure}%
    \vspace{-5pt}
    \caption{UTD32 Returns on Full DMC15-500K.}
    \label{fig:utd32_ret}
\end{figure}

\newpage

\subsection{Q-values on all environments} \label{app:exp_q}

\begin{figure}[H]
\centering
    \begin{subfigure}[b]{0.8\textwidth}
        \centering
        \includegraphics[height=0.8cm]{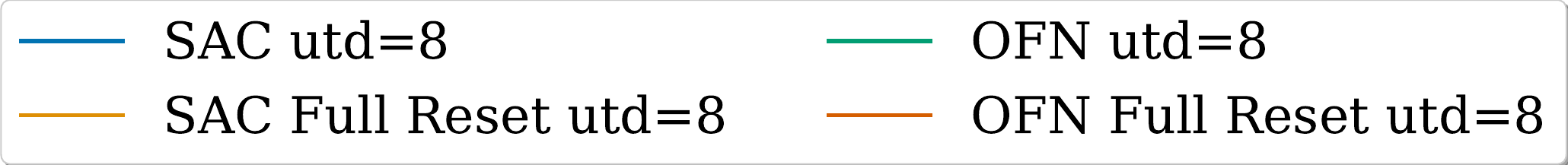}
    \end{subfigure}\\%
    \begin{subfigure}[b]{1\textwidth}
        \centering
        \includegraphics[width=15cm, trim=0cm 0cm 0cm 0cm ,clip]{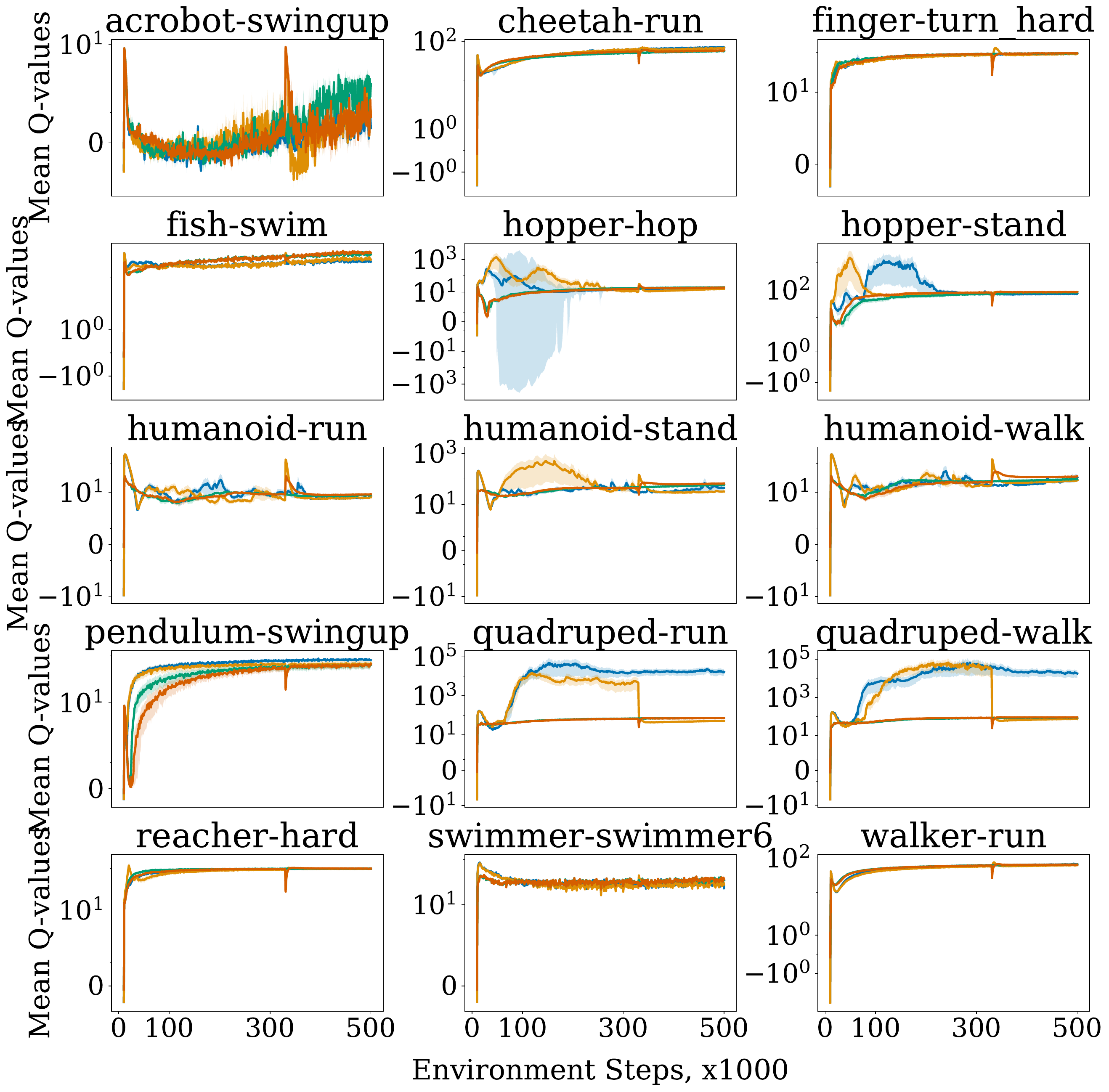}
    \end{subfigure}%
    \vspace{-5pt}
    \caption{UTD8 Q-values on Full DMC15-500K. Resetting often works when Q-values diverge. ONF mitigates divergence.}
    \label{fig:utd8_Q}
\end{figure}

\newpage

\begin{figure}[H]
\centering
    \begin{subfigure}[b]{0.8\textwidth}
        \centering
        \includegraphics[height=1.1cm]{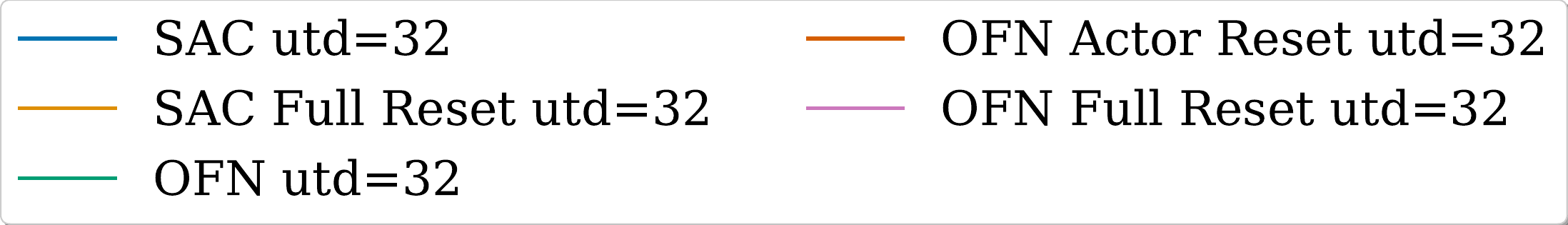}
    \end{subfigure}\\%
    \begin{subfigure}[b]{1\textwidth}
        \centering
        \includegraphics[width=15cm, trim=0cm 0cm 0cm 0cm ,clip]{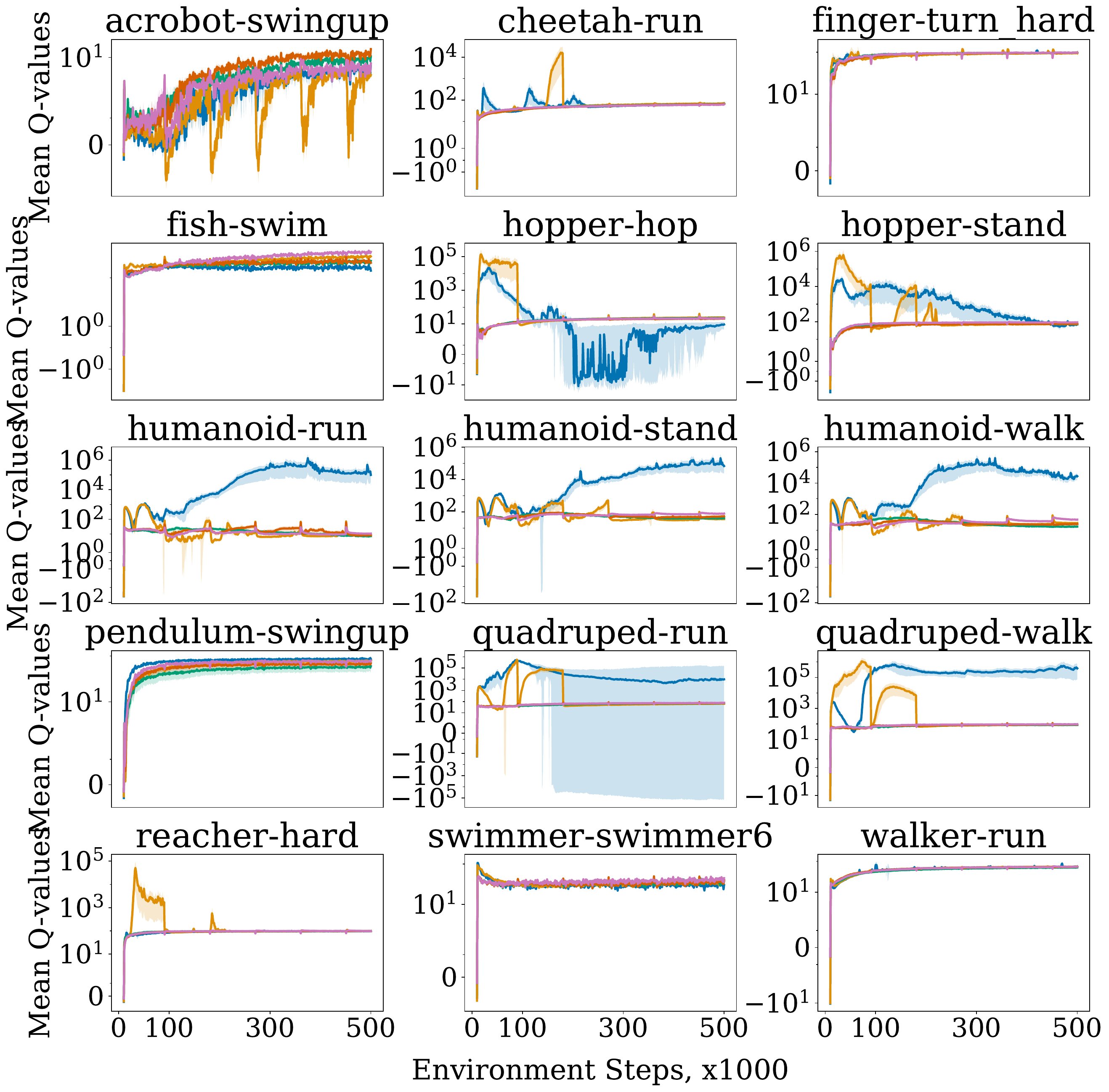}
    \end{subfigure}%
    \vspace{-5pt}
    \caption{UTD32 Q-values on Full DMC15-500K. Resetting often works when Q-values diverge. ONF mitigates divergence.}
    \label{fig:utd32_Q}
\end{figure}

\newpage

\section{Unit norm gradient derivation} \label{app:unitnorm}

Here, we take a look at the gradient of the unit norm projection.

Let $i \in {1, ..., N}$, for all $\mathbf{x} = (x_1, ..., x_n) \in \mathbb{R}^n \setminus \{0\}$. Suppose $f(\mathbf{x}) = \cfrac{\mathbf{x}}{\| \mathbf{x} \|}$. 

Then, 
\begin{align*}
    \partial_i f(\mathbf{x}) 
    &= \cfrac{\| \mathbf{x} \| e_i - \mathbf{x} \partial_i \| \cdot \| (\mathbf{x}) }{\|\mathbf{x} \|^2} \\
    &= \cfrac{\| \mathbf{x} \| e_i - \cfrac{x_i}{\|\mathbf{x} \|} \mathbf{x}}{\|\mathbf{x} \|^2} \\
    &= \cfrac{1}{\|\mathbf{x} \|} e_i - \cfrac{x_i}{\|\mathbf{x} \|^3} \mathbf{x}
\end{align*}

Note that the second term can grow quite large if the norm of $\mathbf{x}$ is relatively small. Despite this fact, we are able to remedy the exploding gradients using unit norm projection, likely because gradients are small when the norm is small.

\section{Open Problems and Limitations} \label{app:open}

Feature divergence without regularization is an important problem that contributes substantially to the issues facing high-UTD learning
However, as our experiments show, there are many additional open problems that introducing normalization does not address.

\textbf{Understanding actor issues}~~The resetting experiments in \autoref{fig:aggregate} highlight that a part of the performance impact of high UTD comes from the actor optimization, not the critic optimization, as resetting the actor can boost performance without changing the critic.
Our work does not address this issue, and to the best of our knowledge there are no specific attempts to investigate the actor optimization process in deep actor-critic reinforcement learning.

{\bf RL Optimizer}~~ As the priming experiments show (\autoref{fig:priming_opt}), the update dynamics introduced by momentum terms in modern optimizers can exacerbate existing overestimation problems. 
\citet{dabney2014adaptive} derives adaptive step-sizes for reinforcement learning from a theoretical perspective, but the resulting optimization rules have not been adapted to Deep Reinforcement Learning to the best of our knowledge.
A recent study by \citet{asadi2023resetting} shows that resetting the optimizer can have some benefit in the DQN setting, where it can be tied to the hard updates of the target Q network.
In addition, \citet{lyle2023understanding} show that optimizers like Adam can lead to reduced plasticity of neural networks.
However, our experiments also highlight that without the accelerated optimization of modern optimizers, convergence of the Q value can be prohibitively slow, highlighting the urgent need for stable and fast optimization in RL.

{\bf Conservative Learning for Online RL}~~ Most current actor-critic methods use some form of pessimistic value estimate to combat the overestimation bias inherent in off-policy Q learning. i.e. via the use of a twinned Q network \citep{fujimoto2018addressing}.
However, this can lead to pessimistic under-exploration \citep{lan2020maxmin}.
To address this, \citet{moskovitz2021tactical} propose to tune the relative impact of pessimistic and optimistic exploration for the environments, while \citet{lee2021sunrise} show that by combining independent critic estimates from ensembles, a UBC like exploration bound can be computed.
These changes could be combined with the mitigation strategies for the feature layer divergence in future work to mitigate the harmful effects of underexploration further.

As our work shows, some of the previous problems with overestimation might not emerge from the bias introduced by off-policy actions, but from the learning dynamics of neural network updates.
This suggests that more work on the exact causes of overestimation might allow us to move beyond the overly pessimistic twinned network minimization trick without needing costly solutions like ensemble methods.

{\bf Tau}~~ The rate of the target network updates is an important hyperparameter in online RL, either through periodic hard copies \citep{mnih2013playing} or the use of a Polyak averaging scheme \citep{lillicrap2016ddpg}.
Updating the network too fast can exacerbate the impact of value divergence, while updating too slowly can delay learning. Preliminary experiments show a relationship between value divergence and target update speed that requires further investigation.

There have also been attempts to accelerate optimization not via the neural network optimization, but through adapting the updates of the target networks \citep{vieillard2020momentum,farahmand2021pid}.
This is an orthogonal direction to the one presented here, and the interplay between target network updates and neural network optimization steps are an important topic for future work.

{\bf Reward Shaping Impact}~~ In several environments, we observe almost no detrimental effects due to high update ratios, while in others the Q-values diverge even without moving beyond one update per sample collected.
A closer inspection suggests that environments in which the initial reward is small and uninformative are much more prone to lead to catastrophic divergence, suggesting a close connection between reward shaping and divergence.
While sparse reward problems have received much attention in the context of exploration, our findings suggests that they also present a challenge for efficient optimization.
Beyond this phenomenon, the interactions between optimization and explorations have been hypothesized to be a strong contributing factor to the good performance of some algorithms \citep{schaul2022phenomenon}, but the role diverging Q-values play in this phenomenon is to the best of our knowledge mostly unexplored.

\end{document}